\documentclass[preprint,12pt]{elsarticle}

\usepackage{amssymb}
\usepackage{amsmath}
\usepackage{url}
\usepackage{cleveref}
\usepackage{upgreek}
\usepackage{algorithm}
\usepackage{algpseudocode}
\usepackage{framed,multirow}

\usepackage{tikz}
\usepackage[T1]{fontenc}
\algnewcommand{\LeftComment}[1]{\Statex \(\triangleright\) #1}
\newcommand*\circled[1]{\tikz[baseline=(char.base)]{
            \node[shape=circle,draw,inner sep=0.5pt] (char) {#1};}}

\journal{Medical Image Analysis}

\begin{document}

\begin{frontmatter}



\title{AI-based association analysis for medical imaging using latent-space geometric confounder correction} 


\author{X. Liu$^{1*}$, B. Li$^{2*}$, M.W. Vernooij$^{1}$, E.B Wolvius$^{1}$, G.V. Roshchupkin$^{1\#}$, E.E. Bron$^{1\#}$} 

\affiliation{organization={$^{1}$Erasmus MC},
            city={Rotterdam},
            country={the Netherlands}}
\affiliation{organization={$^{2}$Harvard Medical School},
            city={Boston},
            country={MA, USA}}
\affiliation{country=$^{*}$These authors contribute equally to this work}
\affiliation{country=$^{\#}$These authors contribute equally to this work}
\begin{abstract}
{This study addresses the challenges of confounding effects and interpretability in artificial-intelligence-based medical image analysis. Whereas existing literature often resolves confounding by removing confounder-related information from latent representations, this strategy risks affecting image reconstruction quality in generative models, thus limiting their applicability in feature visualization. To tackle this, we propose a different strategy that retains confounder-related information in latent representations while finding an alternative confounder-free representation of the image data.

Our approach views the latent space of an autoencoder as a vector space, where imaging-related variables, such as the learning target (t) and confounder (c), have a vector capturing their variability. The confounding problem is addressed by searching a confounder-free vector which is orthogonal to the confounder-related vector but maximally collinear to the target-related vector. To achieve this, we introduce a novel correlation-based loss that not only performs vector searching in the latent space, but also encourages the encoder to generate latent representations linearly correlated with the variables. Subsequently, we interpret the confounder-free representation by sampling and reconstructing images along the confounder-free vector.

The efficacy and flexibility of our proposed method are demonstrated across three applications, accommodating multiple confounders and utilizing diverse image modalities. Results affirm the method's effectiveness in reducing confounder influences, preventing wrong or misleading associations, and offering a unique visual interpretation for in-depth investigations by clinical and epidemiological researchers.
The code is released in the following GitLab repository:}
\\
\url{https://gitlab.com/radiology/compopbio/ai_based_association_analysis}

\end{abstract}


\end{frontmatter}



\section{Introduction}
\label{sec:introduction}
Artificial intelligence (AI) has arisen as a powerful asset across various domains, largely due to its capacity to identify discriminative patterns within high-dimensional data sets. This capacity has been particularly useful in the realm of medical imaging analysis, where AI techniques have proven successful for diagnostic and prognostic prediction tasks \citep{shen2017deep}. However, the application of AI in medical imaging-based association analysis for epidemiological studies has met with certain challenges. These include deriving clinically or epidemiologically significant insights from AI-generated results, a task which, when compared to traditional statistical methods \citep{muggli2017association,howe2019prenatal,roshchupkin2016grey}, has proven difficult \citep{sung2023co,duffy2022confounders}. This is primarily due to two factors: the complexity of visualizing non-linear modeling in AI models, which is also known as the "black box" issue; and the lack of control over confounding variables. These hurdles highlight the need for more interpretable and confounder-free AI models in medical applications.

\subsection{Related work}
\subsubsection{Confounder}
A confounder is a variable that simultaneously affects both the independent and dependent variables in an association analysis \citep{peters2017elements,stewart2022basic}. Consequently, it can falsely create, amplify, or reduce an association between these variables. For example, in a study investigating the association between disease severity (independent variable) and patient recovery time (dependent variable), the patient's overall health status before the onset of the disease could serve as a confounder, since both disease severity and recovery time could be influenced by the patient's overall health status. Therefore, without controlling for this confounder, AI models predicting recovery time might inadvertently factor in features related to the overall health status. And these features in turn will confound the outcome, obscuring the influence of disease severity alone on recovery time. 

Bias in AI often stems from issues such as selection bias or data imbalance \citep{lee2021learning,kim2021biaswap,li2019repair}. It occurs when the data used to train the model is not representative of the broader population, leading to biased results and poor performance. In studies regarding bias, researchers normally assume biased training data and unbiased testing data. They propose methods aimed at improving the prediction performance on the testing data; Different from selection bias or data imbalance, the confounder is generally present in the real-world scenarios which cannot easily be removed by re-selecting or re-sampling of the data. Therefore, in the context of confounder, researchers normally assume that the confounder exists in both the training and testing data. The aim is to detect if there is still association between dependent and independent variables after removing the confounding effects.

Although this confounding problem and its solutions are well-established  in classical epidemiological studies \citep{stewart2022basic,voynov2020unsupervised,pourhoseingholi2012control}, the field of AI provides limited methodology to correct for confounders. Furthermore, this issue becomes exacerbated in scenarios with multiple confounders, making the development of confounder-free AI models essential.

\subsubsection{Confounder Control in AI: Challenges for Medical Imaging}
In the context of AI, topics related to confounder control are variously termed as fair representation learning \citep{pham2023fairness,louizos2015variational,creager2019flexibly,liu2021projection,sarhan2020fairness}, debiased representation learning \citep{lee2021learning,kim2021biaswap}, universal representation learning \citep{li2021universal}, or invariant feature learning \citep{xie2017controllable,akuzawa2020adversarial}. These approaches predominantly focus on learning representations from input data that are related to a specific attribute (i.e., the learning target) while remaining independent of sensitive attributes (i.e., confounders). Especially, when the sensitive attribute refers to different source of domains, these methods overlap with the research topic domain adaptation \citep{pham2023fairness,li2021universal,akuzawa2020adversarial}. 

Whereas the concept of a confounder originates from epidemiology, most methods proposed in the AI field typically have different tasks and goals, and therefore do not fully address the challenges posed by confounders in the context of medical applications \citep{duffy2022confounders,brookhart2010confounding}. 1) Confounders are often continuous variables (e.g., age), yet most existing methods are designed to handle a single binary (e.g., sex) or categorical (e.g., ethnicity) confounder \citep{pham2023fairness,edwards2015censoring,zhang2018mitigating,louizos2015variational,li2021universal,akuzawa2020adversarial,lee2021learning,kim2021biaswap,sarhan2020fairness,zemel2013learning,shen2022fair}. These methods require dividing a batch of training samples into several subgroups (e.g., male and female) to remove confounding effects (e.g., sex), making them unsuitable for continuous confounders, which are more challenging to address; 2) This issue becomes more pronounced when considering the common presence of multiple confounders in medical studies, and only a limited number of existing methods \citep{lu2021metadata,vento2022penalty} can mitigate the joint effects of multiple confounders; 3) In addition, labels (both learning target and confounder) are often missing in medical image analysis, but the exploration of semi-supervised settings that utilize image data with missing labels in confounder-free models remains limited; 4) Existing methods overlook image feature visualization in their design, making it difficult to interpret or understand the findings in image-based association studies.

Recognizing this gap, it is crucial to direct attention towards method development that specifically targets these challenges. For example, CF-Net \citep{zhao2020training}, a confounder-free model based on  adversarial training, replaces the typical cross-entropy or mean squared error (MSE) adversarial loss used in domain adaptation tasks with a statistical correlation-based adversarial loss. This adjustment enhances its support for continuous confounders. Another notable method is MDN \citep{lu2021metadata,vento2022penalty}, which combines linear regression with a unique layer inserted into neural networks. This layer is specifically designed to filter out confounding information, permitting only the residual signals to pass to subsequent layers. Due to the inherent properties of linear regression, this approach supports multiple confounders. Despite these examples, as of today, there remains a scarcity of AI methods that focus on the challenges posed by confounders in the context of medical imaging.

\subsubsection{Challenges of interpretability in confounder-free AI models}
Most related techniques for feature interpretation in medical imaging research field can be grouped into two main categories \citep{van2022explainable,fan2021interpretability}. The first category encompasses gradient and backpropagation methods, which typically examine the gradients or activations within the AI model, creating a saliency map linked to an input image, and highlighting the regions that are most influential to the predicted outcome. Example methods include Gradient-weighted Class Activation Mapping (Grad-CAM) \citep{selvaraju2017grad}, SHAP \citep{lundberg2017unified}, DeepTaylor \citep{montavon2017explaining}, and Layer-wise backpropagation \citep{bach2015pixel}. These methods can be directly attached to a trained confounder-free model. For example, CF-Net \citep{zhao2020training} employs Grad-CAM to generate saliency maps, comparing results with or without confounder control.

Methods in the second category, as referenced in studies \citep{liu2021projection,balakrishnan2021towards,higgins2016beta,stone2017teaching,zhao2019variational}, typically involve customizing a built-in generative model to manipulate the latent space. This manipulation enables the reconstruction of a sequence of images that display image deformations linked to a target attribute, such as age \citep{zhao2019variational}. Typical examples are $\beta$-VAE \citep{higgins2016beta} and infoGAN \citep{chen2016infogan}. Unlike saliency maps, these reconstructed images provide richer semantic information \citep{zhao2019variational}, thereby enhancing the understanding of established associations. However, the interpretability of these methods is intrinsic and not transferable to other models. This limitation necessitates the development of new generative models equipped for confounder control. This task, however, has proven challenging. Removing confounder-related information from the latent space, a common approach in many confounder-free models \citep{edwards2015censoring,xie2017controllable,zhang2018mitigating,zhao2020training,louizos2015variational,alemi2016deep,creager2019flexibly,lu2021metadata,vento2022penalty,shen2022fair}, can adversely affect the quality of image reconstruction \citep{liu2021projection}. This issue arises from the contradiction of two optimization objectives: the image reconstruction aims to keep as much information as possible in the latent space, such as age details in brain image reconstructions, while the confounder mitigation aims to remove all confounder-related information. This contradiction becomes even more pronounced in scenarios involving multiple confounders.

\subsection{Motivation and introduction of our proposed method}
To address above-mentioned challenges and therefore leveraging the potential for semantic feature interpretation in generative models, we explore a different strategy for confounder control without compromising the image reconstruction quality. Recent studies on generative adversarial network (GAN) -based models \citep{voynov2020unsupervised} have revealed most image-related variables (e.g., age in brain image) have a vector direction in the latent space that predominantly captures its variability. Inspired by this, we propose retaining confounder-related information in the latent space, while exploring the identification of a vector direction related to the learning target while being independent of multiple confounders.

We thus introduce a new algorithm for image-based association analysis that not only offers flexibility for correcting multiple confounders but also enables semantic feature interpretation. We consider the latent space of an autoencoder as a vector space, where most imaging-related variables (e.g., a learning target \textbf{t} and a confounder \textbf{c}) have a vector direction that captures their variability. Then the confounding issues are solved by determining a confounder-free vector which is orthogonal to $\vec{c}$ but maximally collinear to $\vec{t}$ (\cref{fig:fig6_1}a). To achieve this, we propose a novel correlation-based loss that not only performs vector searching in the latent space, but also encourages the encoder to generate latent representations linearly correlated with the variables. Afterwards, we interpret the confounder-free representation by sampling and reconstructing images along the confounder-free vector.

\subsection{Distinction from prior work}
In most proposed methods, the correction of confounder is achieved by purging the confounder-related information from learned representations, via adversarial training \citep{edwards2015censoring,xie2017controllable,zhang2018mitigating,zhao2020training,akuzawa2020adversarial}, maximum mean discrepancy (MMD) \citep{louizos2015variational,shen2022fair}, or mutual information (MI) \citep{alemi2016deep,creager2019flexibly} techniques. Generally, these existing techniques have their own limitations in practice. For example, the MMD metric can only handle binary or categorized confounders; The MI is known to be too computationally complex to be included as a loss term, and normally it is required to insert additional neural networks, such as a MI estimator \citep{belghazi2018mutual} or a MI gradient estimator \citep{wen2020mutual}, as an alternative solution; Adversarial training is known to be unstable because it is difficult to balance two competing objectives. To avoid these potential issues, our method adopts a different technique, namely vector orthogonalization, for confounder control. 
Regardless, there are several existing methods highly related to our work. Particularly, the VFAE \citep{louizos2015variational} also utilize autoencoder-based model architecture, and also performs confounder correction in the latent space. However, similar to most existing methods, it attempts to remove all confounder-related information from the latent space, which can highly compromise the image reconstruction quality. In contrast, our method retains confounder-related information in the latent space. Another work \citep{balakrishnan2021towards} might have overlap with our proposed method regarding vector orthogonalization, but their work is fundamentally different from ours. Their work, based on an unsupervised GAN model, lacks an inference path for input images, making it inapplicable to prediction tasks. In their work, a QR-factorization approach is used to solve the vector orthogonalization. This approach firstly estimates a vector $\vec{t}$ for the target and a vector $\vec{c}$ for the confounder, then performs the vector orthogonalization based on them. However, we propose a correlation loss term as a novel solution for vector orthogonalization instead. Joint training the encoder with this loss term is necessary to ensure that the variability of variables is linearly captured in the latent space (\cref{fig:fig6_4}). In contrast, this is not possible with the QR-factorization approach, as it cannot be incorporated into the training of neural networks. In addition, our correlation-based loss term can be easily applied to cases with multiple confounders, which is previously mentioned as a main challenge in medical studies.

\subsection{Summary of contribution}

This study significantly expands upon our previously published conference paper \citep{liu2021projection}. Overall, the main contributions of our work are summarized as follows:

\begin{itemize}

\item To our best knowledge, this is the first to introduce geometric insights of confounders into the latent space of an autoencoder. Subsequently, inspired by geometric interpretation of Pearson’s correlation, we propose a correlation-based loss as a novel solution for confounder correction via vector orthogonalization;
\item Benefitted from the geometric insights, our proposed method 1) can easily handle multiple categorized or continuous confounders, and 2) enables semantic feature interpretation in confounder-free prediction models, facilitating in-depth investigations by clinical and epidemiological researchers;
\item We demonstrate the performance of the proposed approach and its value in three application with either synthetic image or real medical images in a population-based research setting. Experimental results show the potential of our approach as a promising toolset for enhancing association analysis in medical imaging.
\end{itemize}

\begin{figure}[!h]
\centerline{\includegraphics[trim={1.5cm 0cm 0cm 0cm}, clip, width=1.0\linewidth]{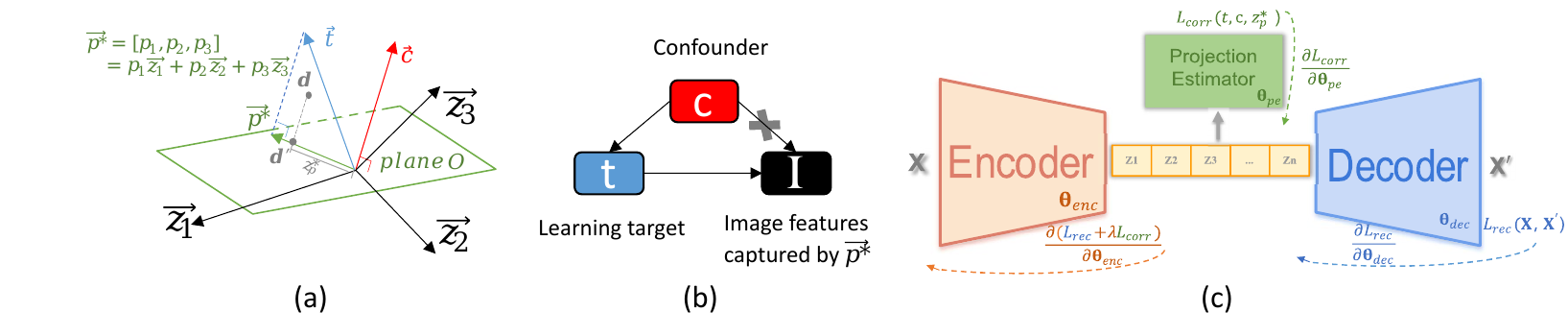}}
\centering
\caption{The proposed AI approach for association analysis in medical imaging. \textbf{(a)} Geometry perspective of correlations between a target and a confounder variable ($\textbf{t}, \textbf{c}$), and its extension ($\vec{t}, \vec{c}$) into the latent space (n=3 latent dimensions) of an autoencoder. Plane O is orthogonal to $\vec{c}$. $\vec{{p}^{*}}$ is the vector projection of $\vec{t}$ onto plane O. $\textbf{d}$ is the latent representation of an input image and $\textbf{d}^{\prime}$ is its projection onto $\vec{{p}^{*}}$. $z_{p}^{*}$ is the distance between $\textbf{d}^{\prime}$ and the origin. For cases with $m$ confounders, the latent dimensions should be $n \geq m + 1$, so as to guarantee there exist a $\vec{{p}^{*}}$ orthogonal to $m$ confounders; \textbf{(b)} a directed acyclic diagram explains the relationships between $\textbf{t}, \textbf{c}$, and image I. We aim to extract image features associated with the learning target while being independent to the confounders. \textbf{(c)} The proposed approach in a neural network perspective. $[z_{1}, z_{2}, ..., z_{n}]$ are the learned latent features by the network, which construct the latent space shown in (a); $\textbf{X}$ and $\textbf{X}^{\prime}$ refer to the input and reconstructed image; $\uptheta_{enc}$, $\uptheta_{dec}$, $\uptheta_{pe}$ are the trainable parameters of encoder, decoder, and projection estimator.}
\label{fig:fig6_1}
\end{figure}

\section{Methodologies}
\subsection{Confounders in a geometric prospective in the latent space}

The latent space of an autoencoder can be considered as a vector subspace of $\mathbb{R}^n$ ($n$ latent dimensions) with basis vectors $\vec{z}_{\text{j}} (j = 1, 2,...,n)$, in which any vector can be represented by $\vec{r}=[r_{1}, r_{2}, ..., r_{n}]$ as a linear combination of the basis vectors (\cref{fig:fig6_1}a). In latent space, most variables (e.g., the learning target, confounders) have a vector direction that predominantly captures its variability \citep{voynov2020unsupervised}. By sampling along this vector, a sequence of images can be reconstructed to explain the association between the input images and the variable, and thus we refer to such vectors as \emph{variable-related vectors}. We assume there exist variable-related vectors $\vec{t}=[t_{1}, t_{2}, ..., t_{n}]$ and $\vec{c}=[c_{1}, c_{2}, ..., c_{n}]$ that represent the learning target $\textbf{t}^{N}$ ($N$ sample size) and a confounder $\textbf{c}^{N}$ in the latent space. The confounding issues in this case would be, there is a correlation between $\vec{t}$ and $\vec{c}$, and therefore, the association that $\vec{t}$ explains between the input images and \textbf{t} is confounded by \textbf{c}.

To mitigate this confounding effect in the latent space, we aim to find an alternative vector $\vec{{p}^{*}}=[p_{1}, p_{2}, ..., p_{n}]$ that is independent to $\vec{c}$ while maximally correlated with $\vec{t}$, such that the vector $\vec{{p}^{*}}$ explains the correlation between the input images and \textbf{t}, without being confounded by \textbf{c}. We refer to $\vec{{p}^{*}}$ as a \emph{confounder-free vectors}. As shown in \cref{fig:fig6_1}a, a $\vec{{p}^{*}}$ independent to $\vec{c}$ can be any vector on the plane O that is orthogonal to $\vec{c}$. With this constraint, the correlation between $\vec{{p}^{*}}$ and $\vec{t}$ reaches the maximum if and only if $\vec{{p}^{*}}$ is collinear with the vector projection of $\vec{t}$ onto the plane O. To solve this geometric orthogonalization problem and to determine $\vec{{p}^{ *}}$ we propose a statistics-based solution in two steps:
\begin{enumerate}
\item Given an input image, its latent representation $\textbf{d}^{n}=(d_{1}, d_{2}, ..., d_{n})$ is computed via the encoder, and shown as a datapoint in the latent space (\cref{fig:fig6_1}a). Subsequently, we calculate $z_{p}^{*}=[d_{1}, d_{2}, ..., d_{n}] \cdot [p_{1}, p_{2}, ..., p_{n}]/\|\vec{{p}^{*}}\|=(p_{1}d_{1}+p_{2}d_{2}+...+p_{n}d_{n})/\|\vec{{p}^{*}}\|$ as the distance between $\textbf{d}^{\prime}$ and the origin, where $\textbf{d}^{\prime}$ is the point projection of \textbf{d} onto $\vec{{p}^{ *}}$. Therefore, $\vec{{p}^{ *}}$ captures the variability of $z_{p}^{*}$. We refer to $z_{p}^{*}$ as a \emph{confounder-free representation} of its input image.
\item To solve $\vec{{p}^{ *}}$, we perform maximum a posteriori estimation using a statistics-based correlation loss term Equation~\eqref{eq:6_3}. The correlation loss iteratively enforces the parameters $[p_{1}, p_{2}, ..., p_{n}]$ to approach the optimal value that minimizes the correlation between $\textbf{z}_{\textbf{p}}^{*N}$ and $\textbf{c}^{N}$, and maximizes the correlation between $\textbf{z}_{\textbf{p}}^{*N}$ and $\textbf{t}^{N}$ over the dataset. Since the Pearson’s correlation coefficient $(r(.))$ equals the cosine value between the two vectors, i.e., $r(\textbf{z}_{\textbf{p}}^{*N}, \textbf{c}^{N})=\cos<\vec{{p}^{*}}, \vec{c}>$ \citep{marks1982note,gniazdowski2013geometric}, the optimization of correlations will drive the search of vector $\vec{{p}^{*}}$. Besides, from \cref{fig:fig6_1}a it can be inferred that there exists an upper bound for the optimization:
\begin{equation}
|r(\textbf{z}_{\textbf{p}}^{*},  \textbf{t})| \leq \sqrt{1-r^{2}(\textbf{t}, \textbf{c})},\hspace{3 mm} s. t. \hspace{2 mm}  r(\textbf{z}_{\textbf{p}}^{*},  \textbf{c})=0
\label{eq:6_1}
\end{equation}
\end{enumerate}

The proposed statistics-based correlation loss has several advantages, compared with an existing QR-factorization approach \citep{balakrishnan2021towards} for vector orthogonalization. An important one is that it not only directly estimates $\vec{{p}^{*}}$ bypassing the need of estimating $\vec{t}$ and $\vec{c}$, but also encourages the encoder to optimally extract latent features that are linearly correlated to the learning target. Without this correlation loss, the linear correlation between latent features and the variables could not be guaranteed. In addition, it can be easily applied to cases with multiple ($m$) confounders Equation~\eqref{eq:6_4}. For cases with $m$ confounders, the latent dimensions should be $n \geq m + 1$, to ensure the existence of a $\vec{{p}^{*}}$ that is simultaneously orthogonal to all $m$ confounders in the latent space (Figure S4 in supplementary).

\subsection{Neural network and loss functions}
We implement the proposed algorithm in an autoencoder framework, consisting of an encoder ($\mathrm{ECN}$), a decoder ($\mathrm{DEC}$), and a projection estimator ($\mathrm{PE}$) (\cref{fig:fig6_1}c). The autoencoder takes $N$ images $\textbf{X}^{N}$ as input, and outputs the reconstructed images $\textbf{X}^{\prime N}$ from the latent representations $\textbf{D}^{N\times{n}}=(\textbf{d}_{\textbf{1}}^{n}, \textbf{d}_{\textbf{2}}^{n}, ..., \textbf{d}_{\textbf{N}}^{n})$. By maximizing the structural and intensity similarities between  $\textbf{X}^{N}$ and $\textbf{X}^{\prime N}$, we expect essential information of $\textbf{X}^{N}$ to be well preserved by $\textbf{D}^{N\times{n}}$ .We optimize this objective using a reconstruction loss term ($L_{rec}$), as measured by the mean absolute error ($\mathrm{MAE}$) between $\textbf{X}^{N}$ and $\textbf{X}^{\prime N}$:
\begin{equation}
L_{rec}(\textbf{X}^{N}, \textbf{X}^{\prime N})=\mathrm{MAE}(\textbf{X}^{N}, \textbf{X}^{\prime N})
\label{eq:6_2}
\end{equation}

Subsequently, we estimate $\vec{{p}^{*}}=[p_{1}, p_{2}, ..., p_{n}]$ in an a-posteriori manner by formulating it into a trainable linear layer within the autoencoder, which is denoted as a projection estimator. The optimal $\vec{{p}^{*}}$ and therefore $\textbf{z}_{\textbf{p}}^{*}$, is found by optimizing the proposed correlation loss term ($L_{corr}$), which is defined as:
\begin{equation}
L_{corr} ( \textbf{t}^N,\textbf{c}^N,\textbf{z}_{\textbf{p}}^{*N})=-|r(\textbf{z}_{\textbf{p}}^{*N},\textbf{t}^N)|+\eta|r(\textbf{z}_{\textbf{p}}^{*N},\textbf{c}^N)|
\label{eq:6_3}
\end{equation}
where $|r(.)|$ is the absolute value of the Pearson correlation coefficient; $\eta$ is a hypterparameter to weight the correlations. An $\eta>>1$ tends to minimize $r(\textbf{z}_{\textbf{p}}^{*N},\textbf{c}^N)$ to 0 and it  enforces a greater correction for confounders, whereas an $\eta=0$ indicates no correction for confounders. This correlation loss is implemented using a mini-batch optimization approach, which has been shown to be both practical and effective for optimizing correlations at the population level \citep{andrew2013deep,zhao2020training,chen2022fine,cao2022pkd}. It handles binary, categorical and continuous confounders. For categorical variables, $\textbf{c}^{N}$ can be converted into dummy variables. Our method can handle higher-order non-linear confounding effects by adding higher-order terms of the confounder (e.g., $\mathrm{Age}^2$) as an additional confounder. For cases with multiple confounders $\textbf{c}_{\textbf{i}} (i=1,2,...,m)$, we extend Equation~\eqref{eq:6_3} and propose  $L_{corr}$ as:
\begin{equation}
\begin{aligned}&L_{corr}\left(\textbf{t},\textbf{c}_{\textbf{1}},\textbf{c}_{\textbf{2}},\ldots,\textbf{c}_{\textbf{m}},\textbf{z}_{\textbf{p}}^{*}\right)=\\&-|r(\textbf{z}_{\textbf{p}}^{*},\textbf{t})|+\eta\left(|r(\textbf{z}_{\textbf{p}}^{*},\textbf{c}_{\textbf{1}})|+\cdots+|r(\textbf{z}_{\textbf{p}}^{*},\textbf{c}_{\textbf{m}})|\right)\end{aligned}
\label{eq:6_4}
\end{equation}

Combining objectives of Equation~\eqref{eq:6_2} and Equation~\eqref{eq:6_3}, we optimize the entire framework using a multi-task loss function ($L_{joint}$):
\begin{equation}
L_{joint}=L_{rec}+\lambda L_{corr}
\label{eq:6_5}
\end{equation}
where $\lambda$ scales the magnitude of the reconstruction and the correlation terms.

In addition, since for medical applications often (learning target and confounding) variables are only available for part of the dataset, we provide a semi-supervised implementation of the loss function to fully exploit the available data. In particular, for each training batch, we update the parameters in two steps: 1) update by $L_{rec}$ based on the missing-label data (half batch), and 2) update by $L_{joint}$ based on the labelled data (half batch) \citep{liu2021projection}.

\begin{algorithm}
\caption{Overall process of our method}
\label{alg:overall_process}
\begin{algorithmic}[1]
\State \textbf{Input:} $\textbf{t}^N , \, \textbf{c}^N, \, \mathbf{X}^{N} $ (N sample size)
\State Initialize model parameters of $\mathrm{ENC}$, $\mathrm{DEC}$, and $\mathrm{PE}$
\State Split $\textbf{t}^N, \textbf{c}^N,  \mathbf{X}^{N}$ into $\textbf{t}_{train}, \textbf{c}_{train},  \mathbf{X}_{train}$ and $\textbf{t}_{test}, \textbf{c}_{test},  \mathbf{X}_{test}$

\hrulefill

\LeftComment{ \% training phase}
\State Optimize $\mathrm{ENC}$, $\mathrm{DEC}$, and $\mathrm{PE}$ via $L_{joint}$ (Eq.5), with $\textbf{t}_{train}, \textbf{c}_{train},$ and $\mathbf{X}_{train}$ as input
\State Obtain $\vec{{p}^{*}}=[p_{1}, p_{2}, ..., p_{n}]$ ($n$ latent dimensions) from $\mathrm{PE}$
\State $\textbf{zp}^{*}_{train} \gets \mathrm{PE}(\mathrm{ENC}(\mathbf{X}_{train}))$
\State Optimize $\mathrm{LR}$ with $\textbf{zp}^{*}_{train}$ and $\textbf{t}_{train}$ as input

\hrulefill

\LeftComment{\% testing phase}
\State $\textbf{zp}^{*}_{test} \gets \mathrm{PE}(\mathrm{ENC}(\mathbf{X}_{test}))$
\State $\hat{\textbf{t}}_{test} \gets \mathrm{LR}(\textbf{zp}^{*}_{test})$
\State Compute $\mathrm{ERR}(\hat{\textbf{t}}_{test}, \textbf{t}_{test}$), where ERR is r-MSE if $t$ is a continuous variable, or AUC for a binary variable
\State Compute $r(\textbf{zp}^{*}_{test}, \textbf{t}_{test})$ and $r(\textbf{zp}^{*}_{test}, \textbf{c}_{test})$

\hrulefill

\LeftComment{ \% semantic feature visualization}
\State $\bar{\mathbf{d}}=(\bar{d}_1,\bar{d}_2,...,\bar{d}_n) \gets \mathrm{mean}(\mathrm{ENC}(\mathbf{X}_{test}))$
\State Generate an arithmetic sequence $\textbf{t}^h = [t_1, t_2, \dots, t_h]$ that covers the range mean$\pm$SD of $\hat{\textbf{t}}_{test}$
\State Obtain $\mathbf{X}^{h'} = [\mathbf{X}_1', \mathbf{X}_2', \dots, \mathbf{X}_h']$ via Algorithm 2, with $\textbf{t}^h$ as input
\State Compute a heatmap $\textbf{H} \gets \mathbf{X}_h' - \mathbf{X}_1'$
\State \textbf{Output:} $\mathrm{ERR}(\hat{\textbf{t}}_{test}, \textbf{t}_{test}$), $r(\textbf{zp}^{*}_{test}, \textbf{t}_{test})$, $r(\textbf{zp}^{*}_{test}, \textbf{c}_{test})$, $\mathbf{X}^{h'}$, and $\textbf{H}$
\end{algorithmic}
\end{algorithm}

\subsection{Semantic feature visualization}
\label{sec:sematic_feature_visualization}
After optimization of the framework, the parameters of the encoder, decoder, and projection estimator are determined; the confounder-free representation ($z_{p}^{*}=\mathrm{PE}(\textbf{d})$) for each given image can be immediately inferred, and used for confounder-free prediction of the learning target ($t$) using logistic or linear regression ($\mathrm{LR}$), i.e., $\hat{t}=\mathrm{LR}(z_{p}^{*})$. To provide insights into the effect of confounder-free prediction on extracting imaging features, we sample along the confounder-free vector $\vec{{p}^{*}}$ and reconstruct a sequence of images ($\textbf{X}_{\textbf{i}}^{\prime}$) that correspond with the predicted target variable $\hat{t}_{i}$. As formulated in Equation~\eqref{eq:6_6} and Equation~\eqref{eq:6_7}, we use the average latent representation $\bar{\mathbf{d}}=(\bar{d}_1,\bar{d}_2,...,\bar{d}_n)$ over the testing set as the reference point for the sampling, and consider $k_{i}$ as a self-defined parameter to control the step and the range of the sampling and to approach the desired target variable:
\begin{equation}
\hat{t}_{i}=\mathrm{LR}(\mathrm{PE}(\bar{\textbf{d}}+k_{i}\vec{{p}^{*}})), i=1,2,...,h, and
\label{eq:6_6}
\end{equation}

\begin{equation}
\textbf{X}_{\textbf{i}}^{\prime}=\mathrm{DEC}(\bar{\textbf{d}}+k_{i}\vec{{p}^{*}})
\label{eq:6_7}
\end{equation}

In the present study, the range of sampling was set to mean ± 3*standard deviation (SD) for Experiment 1, and mean ± SD for Experiment 2 and 3. We set to $h=11$ and thus generated eleven frames of image for feature interpretation. The difference between the first and the last frame was generated as a heatmap for Experiment 2 and 3. Notably, when sampling along $\vec{{p}^{*}}$, the value of ${z}_{{p}}^{*}$ increases, and this change of ${z}_{{p}}^{*}$ is correlated to the changes in the learning target ($\vec{t}$) and the reconstructed image ($\textbf{X}_{\textbf{i}}^{\prime}$), but invariant to the confounder ($\vec{c}$) (\cref{fig:fig6_1}a).

We provide a pseudo-algorithm \textbf{Algorithm~\ref{alg:semantic_feature_visualization}} with scripts for an automatic configuration of $k_{i}$, that allows users to flexibly control the effect of $t$ in the image feature visualization as per their needs. 

\begin{algorithm}
\caption{Semantic feature visualization approach}
\label{alg:semantic_feature_visualization}
\begin{algorithmic}[1]
\State \textbf{Input:} $t_i$
\State \textbf{Required:} $\mathrm{DEC}, \vec{{p}^{*}}, \bar{\textbf{d}}$, and $\mathrm{LR}$ obtained from Algorithm 1
\State Initialize $k_i \gets 0$
\State $\hat{t}_i \gets \mathrm{LR}(\bar{\textbf{d}} + k_i \vec{{p}^{*}})$
\While {$\frac{|t_i - \hat{t}_i|}{t_i} > 0.001$} \Comment{\% loop while the error $>$ 0.1\%}
    \If {$t_i - \hat{t}_i > 0$}
        \State Slightly increase the value of $k_i$
    \Else
        \State Slightly decrease the value of $k_i$
    \EndIf
    \State $\hat{t}_i \gets \mathrm{LR}(\bar{\textbf{d}} + k_i \vec{{p}^{*}})$
\EndWhile
\State $\mathbf{X}_i' \gets \mathrm{DEC}(\bar{\textbf{d}} + k_i \vec{{p}^{*}})$
\State \textbf{Output:} $k_i, \mathbf{X}_i'$
\end{algorithmic}
\end{algorithm}

\section{Experiments and results}We demonstrate the performance of the proposed approach in three applications using 2D synthetic, 3D facial mesh, and 3D brain imaging data, and showing the use of 2D convolutional autoencoder \citep{hou2017deep}, 3D graph convolutional autoencoder \citep{gong2019spiralNet}, and 3D convolutional autoencoder \citep{li2022ahigh} within our architecture-agnostic framework. In the analysis of 3D brain, we used an autoencoder additionally integrated with a normalized cross-correlation (NCC) as reconstruction loss term. NCC is a widely used metric for evaluating local structural correspondence in medical images due to its robustness against intensity variations \citep{klein2009evaluation}. In our case, NCC complements the voxel-wise similarity captured by the $L_{1}$ norm by emphasizing regional patterns and preserving local anatomical structures. Therefore, the model is encouraged to achieve both precise voxel-wise reconstruction and smooth alignment of local structures, improving the overall reconstruction quality and robustness to intensity inconsistencies. 

For all three Experiments, we applied 5-fold cross-validation and ensured that repeated scans from the same subject were in the same training or testing set. In the first fold, the training sample (80\%) were further split into a training set (70\%) and validating set (10\%) for the tuning of hyperparameters. Specifically, we prioritize the correlation loss term when adjusting the value of $\lambda$ Equation~\eqref{eq:6_5}. We suggest a larger batch size $(>8)$ since the correlation was computed on a batch level Equation~\eqref{eq:6_3}. To successfully mitigate the confounding effect, we suggest an $\eta=2$ Equation~\eqref{eq:6_3}. An $\eta=0$ indicates the proposed method without correction for confounders. By comparing the results of $\eta=2$ and $\eta=0$, we compared the difference between with and without correction for confounders. To better distinguish them, we refer $\vec{p}$ and $z_{p}$ to the results without correction for confounders ($\eta=0$), while referring $\vec{{p}^{*}}$ and $z_{p}^{*}$ to those with correction ($\eta=2$). The latent dimensions was 2, 64 and 64 for Experiment 1, 2 and 3. The batch size was 16, 64, and 8 for Experiment 1, 2 and 3. Epochs were 300 for all Experiments.

In Experiment 1, we conducted method comparison among the following models:
\begin{itemize}
\item Variational autoencoder (VAE): An unsupervised model for 2D image reconstruction, serving as a baseline reconstruction method without any confounder restriction on latent feature.
\item Ours (NA): Our proposed model implemented in a VAE without confounder control, by setting $\eta=0$ in Equation~\eqref{eq:6_3}.
\item Ours (*): Our proposed model implemented in a VAE with confounder control, by setting $\eta=2$ in Equation~\eqref{eq:6_3}.
\item VFAE-MI \citep{louizos2015variational}: A supervised VAE-based confounder-free method, which removes confounder-related information from the latent space via a MMD loss. Since the MMD loss is not applicable to continuous confounders, it is replaced by a MI loss \citep{belghazi2018mutual}.
\item CF-Net \citep{zhao2020training}: A supervised confounder-free deep learning method, which removes confounder-related information from the latent space via adversarial training techniques.
\item PMDN \citep{vento2022penalty}: A supervised deep learning method, which combines linear regression with a unique layer inserted into neural networks. This layer filters out confounding information, permitting only the confounder-free residual signals to subsequent layers.
\end{itemize}

In all experiments, we quantified the performance of prediction Equation~\eqref{eq:6_6} accuracy by the root mean square error (r-MSE) for continuous variables, and by the area under the receiver
operating characteristic curve (AUC) for binary variables, the image reconstruction quality by the mean $L_{1}$-norm between the input and reconstructed images, and the confounding correction by the Pearson’s correlation coefficient between the latent image representation ($z_{p}$) and variables. We also included mutual information (MI) \citep{alemi2016deep} and squared distance correlation ($\mathrm{dcor}^2$) \citep{wikipedia2024distancecorrelation} as additional metrics for the evaluation of confounder removal. These metrics measure both linear and non-linear dependency. A lower $\mathrm{dcor}^2$ or a lower MI reflects lower dependency.

\subsection{Data sets}

\subsubsection{Simulated 2D solid circles}
\label{sec:synthetic_data_set}

We simulated a dataset of 2D images (N=8,000; image size 64x64) with black background and greyscale solid circles of different brightness and radius. To simulate the research scenario of one learning target variable and one confounder variable, we used the brightness of the circle as the learning target and radius as confounder \citep{gitlab2023compopbio}. The brightness (range: 0-1 rescaled from 128-255 in grayscale; mean: 0.470±0.135) and radius (range: 3-30 pixels; mean: 16±6.320 pixels) followed a multivariate Gaussian distribution, and the Pearson correlation coefficient between them is -0.668. The center of all circles lies in the geometric center of the image. A brightness of 0 indicates gray and a brightness of 1 indicates white.

To further demonstrate the effectiveness of the proposed method in scenarios involving multiple confounders, we extended this synthetic dataset to incorporate additional confounding variables. Comprehensive details of the extended dataset and the corresponding experimental results are provided in the supplementary materials.
\subsubsection{Facial shape dataset}
3D facial shape imaging data from the multi-ethnic population-based Generation R cohort study is included (mean age: 9.8 years; N=5,011) \citep{jaddoe2006generation}. The raw facial shape data were acquired from a 3dMD (Atlanta (GA), USA) camera system. We adopted a mesh template representing an average face with a fixed topology, and built the raw data into a mesh-template based data set by deforming the template via NICP registration. \citep{amberg2007optimal, liu2023association}. Thus, all 3D facial images used in the 3D graph autoencoderbshare the same resolution (5,023 vertices) and the same topology (9,844, triangular faces with the same edge connectivity). Additionally, phenotyping regarding sex, BMI, ethnicity, low to moderate prenatal alcohol exposure (PAE) during pregnancy, maternal age, and maternal smoking during pregnancy was performed. The binary phenotypes were digitized, namely, Sex: 1 for female and 0 for male; Ethnicity: 1 for Western and 0 for non-Western; PAE: 1 for exposed and 0 for non-exposed; Maternal smoking: 1 for smoking and 0 for non-smoking. We have access to N=1,515 labelled samples (760 non-exposed and 755 exposed) and N=3,496 missing-label samples. Details about data characteristic are shown in \cref{tab:tab6_2}.

\subsubsection{Aging brain dataset}
\label{sec:aging_brain_dataset}
The Rotterdam Study is a prospective population-based study targeting causes and consequences of age-related diseases among 14,926 participants \citep{ikram2020objectives}. In this work, we included 11,801 3D T1-weighted brain MRI scans from N=5,717 participants (mean age: 64.7±9.8 years, female: 54.5\%) who have no prevalent dementia or stroke at time of MRI. Scans were acquired on one 1.5T MRI scanner (GE Signa Excite; GE Healthcare, Madison, USA). For 3D T1-weighted images, the acquisition parameters were: TR/TE=13.8ms/2.8ms; imaging matrix of 416x256 in an FOV of 250x250mm$^{2}$ \citep{ikram2011rotterdam}. The voxel size was 0.5x0.5x0.8mm$^{3}$. We computed modulated grey matter (GM) maps using voxel-based morphometry (VBM) that describe the local GM density \citep{good2001voxel,roshchupkin2016fine}. The matrix size of the resulting modulated GM maps, that we use as input images for the association analysis, is 160x192x144. As the learning target, we use a general cognitive factor (g-factor) that was computed using principal component analysis incorporating color-word interference subtask of the Stroop test, LDST, verbal fluency test, delayed recall score of the 15-WLT, DOT and Purdue pegboard test \citep{hoogendam2014patterns}. Scans acquired at baseline time-point and with available phenotypes and valid cognitive test results were considered as labeled samples (N=2,395). Additionally, we have access to N=9,406 missing-label samples.

\subsection{Results}

\subsubsection{Results on the synthetic data set}

In the first experiment, we construct a synthetic dataset to facilitate the understanding of the confounding problem as well as the proposed solution. The dataset (Section~\ref{sec:synthetic_data_set}, \cref{fig:fig6_3}) consists of 2D grayscale images of solid circles with different brightness and radius, in which a larger radius is strongly correlated with a lower brightness. Brightness is used as the learning target and the circle radius is used as a confounder. Given that the input images can be represented by two features (brightness and radius), a latent space with two dimensions would be sufficient to fully reconstruct the input images (\cref{fig:fig6_3}). We thus set the latent dimension of this autoencoder to be two.

\begin{figure}[!h]
\centerline{\includegraphics[trim={0.0cm 0.3cm 0cm 0.3cm}, clip, width=0.95\linewidth]{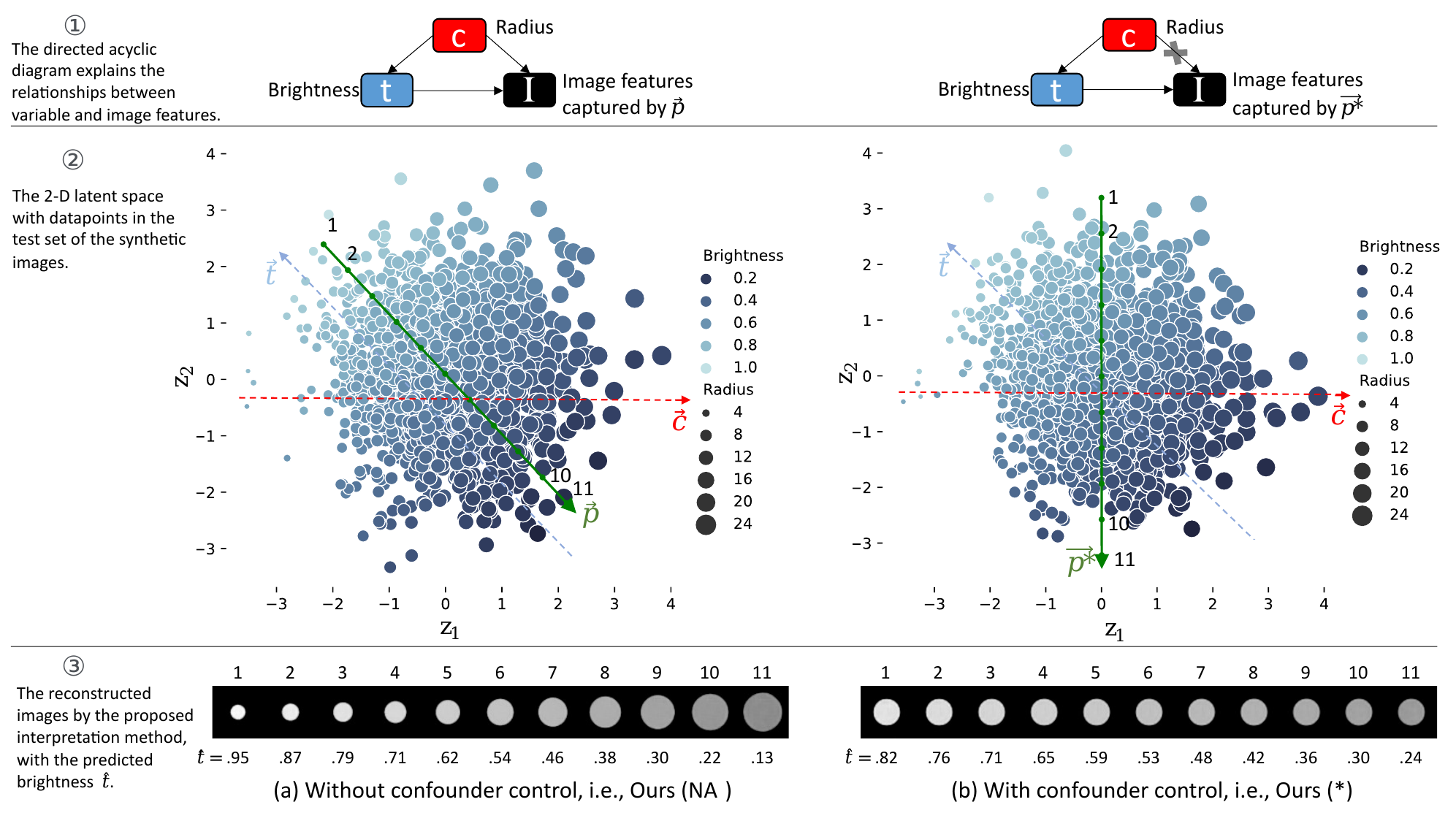}}
\centering
\caption{The distribution of the 2-D latent space for the synthetic images in the test set of Experiment 1, and the eleven reconstructed images sampling along the brightness-related vector, \textbf{(a)} without (i.e., vector $\vec{p}$) and \textbf{(b)} with correction (vector $\vec{{p}^{*}}$) for the confounding of circle radius, together with the predicted brightness $\hat{t}$ derived by Equation~\eqref{eq:6_6} and Equation~\eqref{eq:6_7}. $Z_{1}$-axis: the first dimension of the latent space; $Z_{2}$-axis: the second dimension. Each data point in the latent space represents an input image, which is denoted by its radius and brightness. After training, eleven frames were reconstructed by sampling eleven points along the vector $\vec{p}$ and $\vec{{p}^{*}}$ (Equation~\eqref{eq:6_6} and Equation~\eqref{eq:6_7}) to visualize the confounding effects. Whereas our method does not involve the estimation of vectors $\vec{t}$ and $\vec{c}$, we have manually included them in this figure only for the purpose of enhancing comprehension.}
\label{fig:fig6_2}
\end{figure}

\begin{figure}[!h]
\centerline{\includegraphics[width=1.0\linewidth]{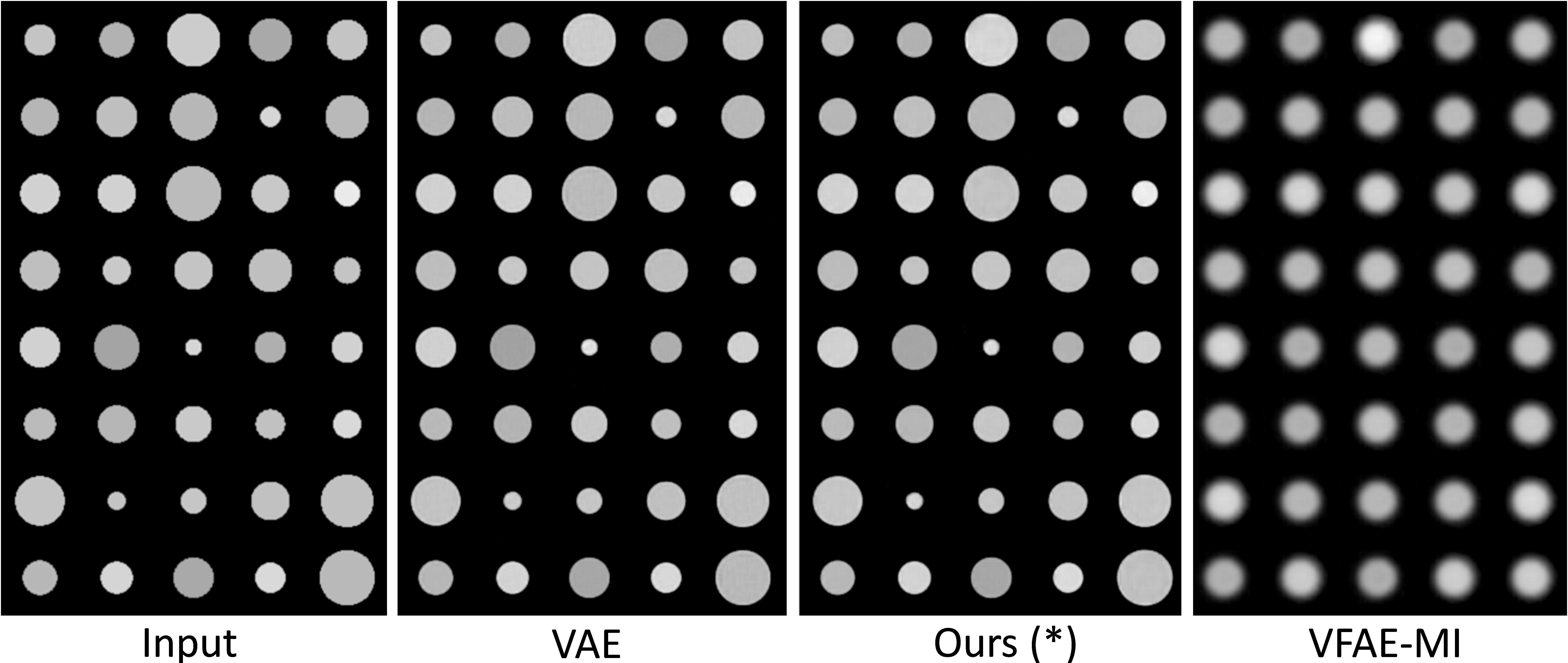}}
\centering
\caption{The input images $\textbf{X}$ (8$\times$5 circle images), and the reconstrued images $\textbf{X}^{\prime}$ of different methods, in Experiment 1.}
\label{fig:fig6_3}
\end{figure}

\begin{figure}[!h]
\centerline{\includegraphics[trim={0.0cm 0.6cm 0.0cm 1.1cm}, clip, width=0.90\linewidth]{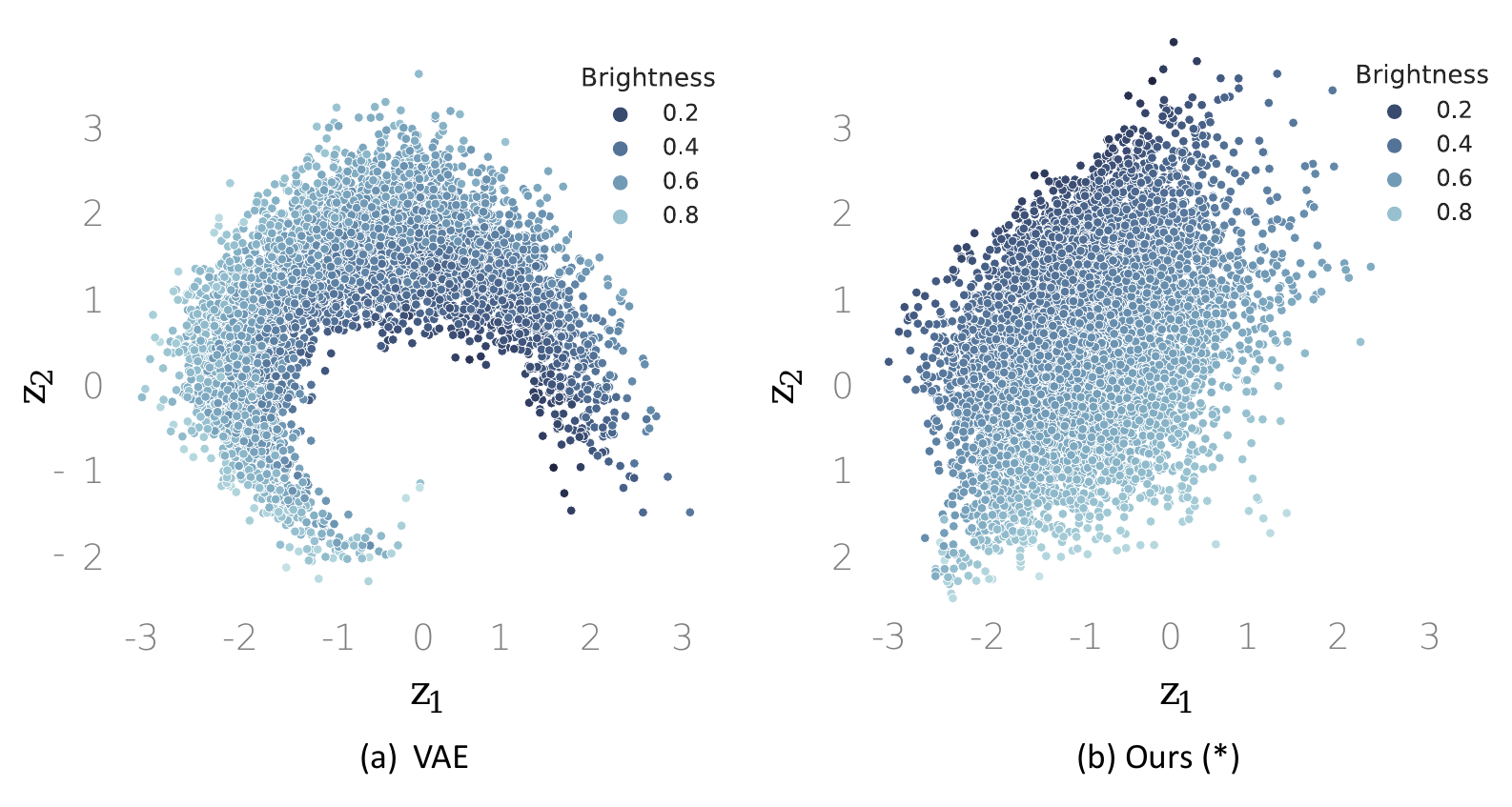}}
\centering
\caption{Distribution of datapoints in the 2-D latent space via \textbf{(a)} unsupervised training and \textbf{(b)} our proposed supervised training, in Experiment 1.}
\label{fig:fig6_4}
\end{figure}

We plot all 2D latent representations of the test set images, together with the estimated brightness-related vector (confounder-free vector $\vec{{p}^{*}}$, confounder-affected vector $\vec{p}$; \cref{fig:fig6_2}). \cref{fig:fig6_2} shows that the variability of both brightness and radius variable are linearly encoded in the latent space following the supervision of the correlation loss Equation~\eqref{eq:6_5}. Please note, without this correlation loss, the linear correlation between latent features and the variables could not be guaranteed (\cref{fig:fig6_4}).

Without adjusting for the confounding of circle radius (\cref{fig:fig6_2}a), vector $\vec{p}$ effectively captured the variation in the circle brightness as indicated by the high correlation of $r(\textbf{z}_{\textbf{p}}, \mathrm{brightness})=-0.992$ (\cref{tab:tab6_1}), demonstrating exceptional prediction accuracy. However, sampling along vector $\vec{p}$ in the 2D latent space not only moved towards data points with a lower brightness but also those with a larger radius (see \cref{fig:fig6_2}a\circled{2}), suggesting a prominent confounding problem. This confounding issue was similarly revealed by the proposed feature interpretation method. When we reconstruct images with decreasing brightness based on the identified association, there was a noticeable increase in the circle's radius (\cref{fig:fig6_2}a $\small{\circled{3}}$), because vector $\vec{p}$ also partially captured the variability of the radius ($r(\textbf{z}_{\textbf{p}}, \mathrm{radius})=+0.668$ in \cref{tab:tab6_1}).

On the contrary, when employing the proposed confounder correction, vector $\vec{{p}^{*}}$ still captured the variance of brightness, but remained almost unaffected by the change in the radius (\cref{tab:tab6_1}; $r(\textbf{z}_{\textbf{p}}^{*}, \mathrm{brightness})=-0.743$, $r(\textbf{z}_{\textbf{p}}^{*}, \mathrm{radius})=+0.013$). Sampling along vector $\vec{{p}^{*}}$ in the latent space led to data points with a lower brightness, but there were no discernible changes in the radius (\cref{fig:fig6_2}b\circled{2}). Similarly, the proposed interpretation method showed that when images were reconstructed with decreasing brightness based on the established confounder-free association, the circle radius visually remained constant (\cref{fig:fig6_2}b$\circled{3}$).

\begin{table}[!h]
\centering
\begingroup
\tiny 
\caption{Prediction error, Pearson’s correlation coefficient, and image reconstruction quality of methods without (NA) and with (*) correction for the circle radius (confounder) in predicting the circle brightness (learning target) on the test set.}
\label{tab:tab6_1}
\setlength{\tabcolsep}{3pt}
\begin{tabular}{l|ll|lll|l}
\hline
                      & \multicolumn{2}{l|}{Prediction performance}     &  \multicolumn{3}{l|}{Confounder removal} & {Image reconstruction} \\ \cline{2-7}
Methods & r-MSE (brightness)   &$r(\textbf{z}_{\textbf{p}}, \mathrm{brightness})$        & $r(\textbf{z}_{\textbf{p}}, \mathrm{radius})$     & $\mathrm{dcor}^2$$(\textbf{z}_{\textbf{p}}, \mathrm{radius})$    & MI$(\textbf{z}_{\textbf{p}}, \mathrm{radius})$       & $L_{1}$-norm              \\ \hline
VAE (NA)                    & -          & -               & -   & -   & -  & \textbf{0.007±0.000}          \\
Ours (NA)                    & \textbf{0.020±0.003} &   \textbf{-0.992±0.001}      & +0.668±0.018        & 0.378±0.018    & 0.304±0.033   & 0.008±0.000 \\
VFAE-MI (*)                     & 0.106±0.026        & -0.729±0.018              & +0.036±0.021  & 0.050±0.034  & 0.051±0.034  & 0.062±0.010          \\
CF-Net (*)                      &0.097±0.007          & -0.732±0.010               & +0.025±0.014   & 0.018±0.004  & 0.076±0.032  & -          \\
PMDN (*)                      & 0.132±0.009          & -0.612±0.057               & +0.071±0.028   & 0.093±0.024  & 0.395±0.183  & -          \\
Ours (*)                     & 0.092±0.003          & -0.743±0.008               & \textbf{+0.013±0.011}   & \textbf{0.001±0.001}  & \textbf{0.024±0.019}  & 0.008±0.000          \\ \hline
\multicolumn{7}{p{380pt}}{+ and - indicates a positive or negative correlation between $\textbf{z}_{\textbf{p}}$ and the variables.}\\
\multicolumn{7}{p{380pt}}{Notion $z_{p}$ in this study is referring to notion $y$ in other methods.}\\
\multicolumn{7}{p{380pt}}{A better performance is indicated by a \textbf{bold} value.}
\end{tabular}
\endgroup
\end{table}

In this synthetic dataset, the Pearson's correlation between circle radius and brightness $r(\textbf{t}, \textbf{c})$ is 0.668 (Section~\ref{sec:synthetic_data_set}). Theoretically, after fully adjusting for the confounder, the upper bound (as defined in Equation~\eqref{eq:6_1}) of any remaining correlation between $\textbf{z}_{\textbf{p}}^{*}$ and the learning target should be $|r(\textbf{z}_{\textbf{p}}^{*},  \textbf{t})| \leq \sqrt{1-r^{2}(\textbf{t}, \textbf{c})}=0.744$, subject to $r(\textbf{z}_{\textbf{p}}^{*},  \textbf{c})=0$. This theoretical upper bound is in line with our observed result of $|r(\textbf{z}_{\textbf{p}}^{*}, \mathrm{brightness})|=0.743$, with $r(\textbf{z}_{\textbf{p}}^{*}, \mathrm{radius})=0.013$ (\cref{tab:tab6_1}). Furthermore, when we compared our approach with other confounder-free methods, we found that all methods obtained similar outcomes in terms of prediction accuracy and confounder control (\cref{tab:tab6_1}). However, in terms of image reconstruction, our method obtained a better performance than the VFAE-MI (\cref{tab:tab6_1}). When VFAE-MI successfully corrected for the confounder, the the radius-related details were removed, resulting in blurred reconstructed images and low image reconstruction quality(\cref{fig:fig6_3}). In contrast, our method did not compromise the image reconstruction quality.

\subsubsection{Results on alcohol exposure prediction from 3D facial shape in children population}
\label{sec:result_facial_shape}
High levels of prenatal alcohol exposure (PAE) during pregnancy can have significant adverse effects on a child's health development resulting in fetal alcohol spectrum disorder (FASD) with abnormal facial development \citep{jones1973recognition}. The association of low–moderate levels of PAE with the child’s facial development is less known. We thus applied the proposed method with multiple-confounder correction to study the associations between low-to-moderate prenatal alcohol exposure and children’s facial shape on a large population-based birth cohort (Generation R, \citep{jaddoe2006generation}. We used maternal alcohol consumption as the learning target (PAE 1: drinking during pregnancy; PAE 0: not drinking), and as suggested by a previously similar study \citep{muggli2017association,howe2019prenatal} using traditional statistical method with traditional confounder control, we used sex, ethnicity, BMI, maternal age, and maternal smoking during pregnancy as confounders, resulting in N=755 exposed and N=760 non-exposed samples. In addition, we included N=3,496 missing-label samples in semi-supervised learning (SSL) setting. The data characteristic is provided in \cref{tab:tab6_2}, where maternal smoking ($p<0.001$, two-sample t-test), maternal age ($p<0.00$1), child BMI ($p<0.001$), and especially ethnicity ($ p<0.001$) showed imbalanced distribution between the non-exposed and exposed groups.

\begin{table}[!h]
\centering
\caption{Data characteristic of children and their mothers included in the analysis (for the labeled data only, N=1,515).}
\label{tab:tab6_2}
\small
\setlength{\tabcolsep}{3pt}
\begin{tabular}{llll}
\hline
Characteristic              & Non-exposed  & Exposed      & Two sample         \\
                            & (N=760)      & (N=755)      & t-test, p-value                   \\ \hline
Child’s ethnicity, No. (\%) &              &              &                           \\
\quad \quad1: Western                  & 328 (43.2\%) & 670 (88.7\%) & \multirow{2}{*}{1.94e-87} \\
\quad \quad0: Non-Western              & 432 (56.8\%) & 85 (11.3\%)  &                           \\
Child’s sex, No. (\%)       &              &              &                           \\
\quad \quad0: Male                     & 357 (47.0\%) & 370 (49.0\%) & \multirow{2}{*}{0.44}     \\
\quad \quad1: Female                   & 403 (53.0\%) & 385 (51.0\%) &                           \\
Child’s BMI, mean ± SD      & 18.6 ± 3.2   & 16.8 ± 2.0   & 4.58e-37                  \\
Maternal smoking, No. (\%)  &              &              &                           \\
\quad \quad1: Yes                      & 204 (26.8\%) & 417 (55.2\%) & \multirow{2}{*}{2.53e-30} \\
\quad \quad0: No                       & 556 (73.2\%) & 338 (44.8\%) &                           \\
Maternal age, mean ± SD     & 28.2 ± 5.0   & 32.1 ± 3.9   & 8.60e-62                  \\ \hline
\end{tabular}
\end{table}

\begin{table}[!h]
\centering
\caption{Association analysis between PAE (learning target) and children’s facial shape (input image). Results are presented without (NA) and with (*) controlling of the confounders (ethnicity, bmi, sex, maternal smoking, maternal age).}
\label{tab:tab6_3}
\setlength{\tabcolsep}{3pt}
\small 
\begin{tabular}{l|llll}
\hline
 & Metrics & Ours (NA) & Ours (*) & Ours-SSL (*) \\ \hline
\multirow{2}{*}{\begin{tabular}[c]{@{}l@{}}Prediction \\ performance\end{tabular}} & AUC (PAE) & \textbf{0.73±0.03} & 0.58±0.03 & 0.59±0.01 \\
 & r(PAE) & \textbf{+0.40±0.06} & +0.12±0.02 & +0.15±0.01 \\ \hline
\multirow{5}{*}{\begin{tabular}[c]{@{}l@{}}Confounder\\ removal\end{tabular}} & $r$(Eth) & +0.31±0.05 & +0.04±0.02 & \textbf{+0.04±0.02} \\
 & $r$(BMI) & -0.33±0.06 & -0.03±0.02 & \textbf{-0.02±0.02} \\
 & $r$(Sex) & +0.05±0.05 & +0.04±0.04 & \textbf{+0.04±0.03} \\
 & $r$(MS) & +0.10±0.06 & \textbf{+0.03±0.03} & +0.04±0.03 \\
 & $r$(MA) & +0.23±0.06 & \textbf{+0.03±0.02} & +0.03±0.03 \\ \hline
\multirow{5}{*}{\begin{tabular}[c]{@{}l@{}}Confounder \\ removal\end{tabular}} & $\mathrm{dcor}^2$(Eth) & 0.12±0.03 & 0.01±0.02 & \textbf{0.01±0.02} \\
 & $\mathrm{dcor}^2$(BMI) & 0.11±0.03 & 0.02±0.01 & \textbf{0.01±0.01} \\
 & $\mathrm{dcor}^2$(Sex) & 0.01±0.01 & 0.01±0.01 & \textbf{0.01±0.00} \\
 & $\mathrm{dcor}^2$(MS) & 0.03±0.01 & \textbf{0.00±0.00} & 0.00±0.00 \\
 & $\mathrm{dcor}^2$(MA) & 0.06±0.03 & 0.00±0.00 & \textbf{0.00±0.00} \\ \hline
\multirow{5}{*}{\begin{tabular}[c]{@{}l@{}}Confounder\\ removal\end{tabular}} & MI(Eth) & 0.10±0.03 & 0.02±0.01 & \textbf{0.01±0.01} \\
 & MI(BMI) & 0.09±0.02 & 0.03±0.02 & \textbf{0.02±0.02} \\
 & MI(Sex) & 0.02±0.02 & 0.01±0.02 & \textbf{0.01±0.01} \\
 & MI(MS) & 0.03±0.02 & \textbf{0.01±0.01} & 0.02±0.01 \\
 & MI(MA) & 0.13±0.03 & 0.05±0.01 & \textbf{0.04±0.01} \\ \hline
\begin{tabular}[c]{@{}l@{}}Image\\ reconstruction\end{tabular} & $L_{1}$-norm & 0.33±0.01 & 0.34±0.01 & \textbf{0.32±0.01} \\ \hline
\multicolumn{5}{p{380pt}}{Eth: Ethnicity of the child}\\
\multicolumn{5}{p{380pt}}{MS: Maternal smoking durning pregnancy}\\
\multicolumn{5}{p{380pt}}{MA: Maternal age}\\
\multicolumn{5}{p{380pt}}{Ours-SSL (*) means our method with confounder control in a semi-supervised setting utilizing additional missing-label training data.}\\
\multicolumn{5}{p{380pt}}{+ and - indicates a positive or negative correlation between $\textbf{z}_{\textbf{p}}$ and the variables.}\\
\multicolumn{5}{p{380pt}}{A better performance is indicated by a \textbf{bold} value.}
\end{tabular}
\end{table}

As results shown in \cref{tab:tab6_3}, when without correction for confounders, the Pearson’s correlation coefficient between the learned facial image features ($z_{p}$) and PAE was +0.40, and the prediction AUC of PAE was at 0.73. The observed correlation coefficients between $z_{p}$ with ethnicity and BMI were +0.31 and -0.33, i.e., at a similar strength as that of the learning target PAE. The observed correlation coefficients for child sex, maternal smoking, and maternal age were +0.05, +0.10 and +0.23. After correcting for the confounders of ethnicity, BMI, sex, maternal smoking, and maternal age using the proposed method, the correlation coefficient between facial image features ($z_{p}^{*}$) and PAE decreased to +0.12, which is higher than the correlation coefficient for the confounders (ethnicity: +0.04; BMI: -0.03; sex: +0.04; maternal smoking: +0.03; maternal age: +0.03). In the SSL setting where additionally missing-label training data were included, we observed improved prediction accuracy and image reconstruction quality, when controlling confounders at a similar level.

\begin{figure}[!h]
\centerline{\includegraphics[width=.7\linewidth]{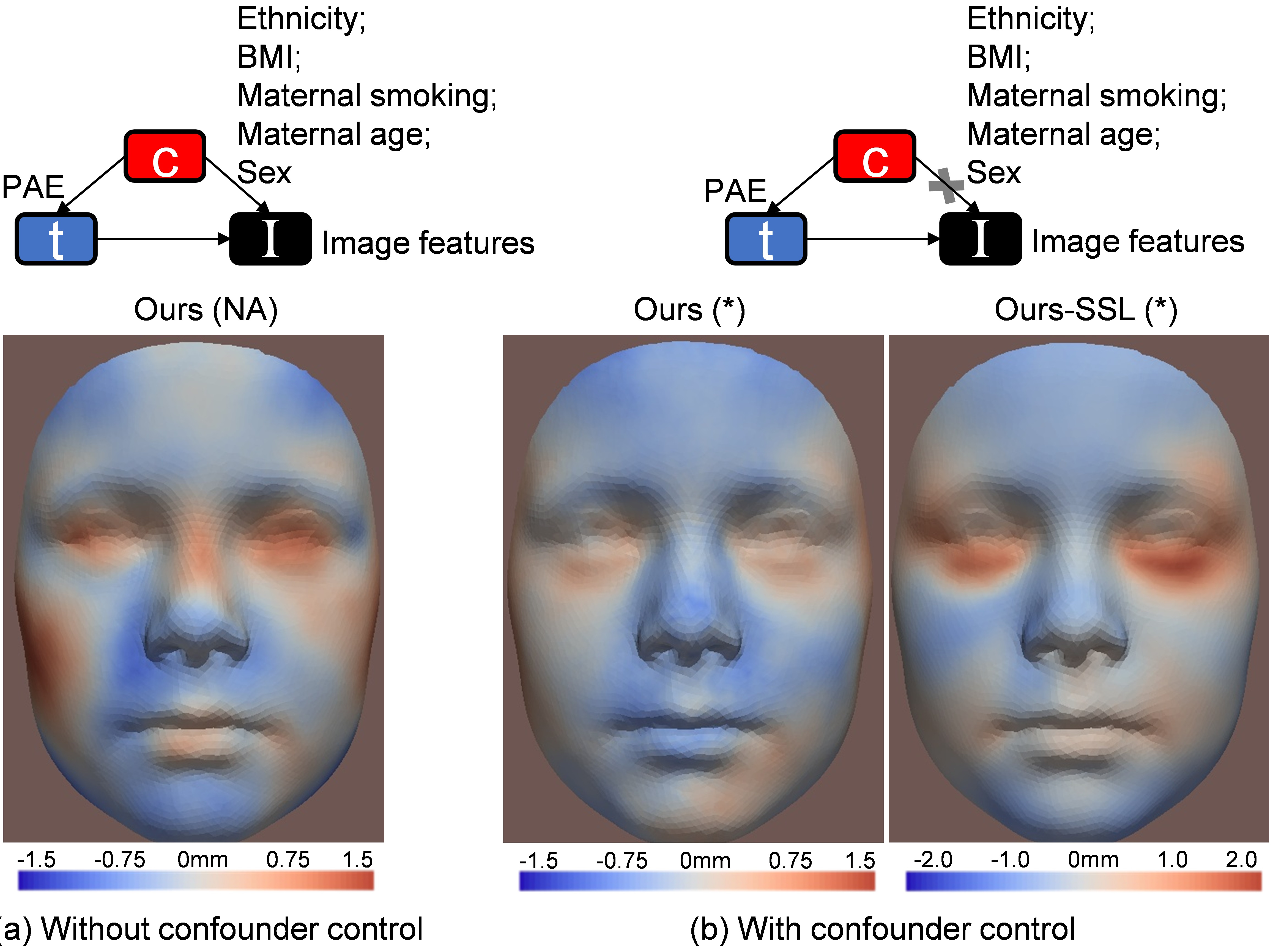}}
\centering
\caption{Interpretation heatmaps of facial changes in children with PAE using the proposed method: \textbf{(a)} without correction for confounders; \textbf{(b)} with correction for ethnicity, BMI, maternal smoking, maternal age, and sex. Red areas refer to inward changes of the face with respect to the geometric center of the head. Heatmap generation is detailed in Section~\ref{sec:sematic_feature_visualization}.}
\label{fig:fig6_5}
\end{figure}

\begin{figure}[!h]
\centerline{\includegraphics[width=1.0\linewidth]{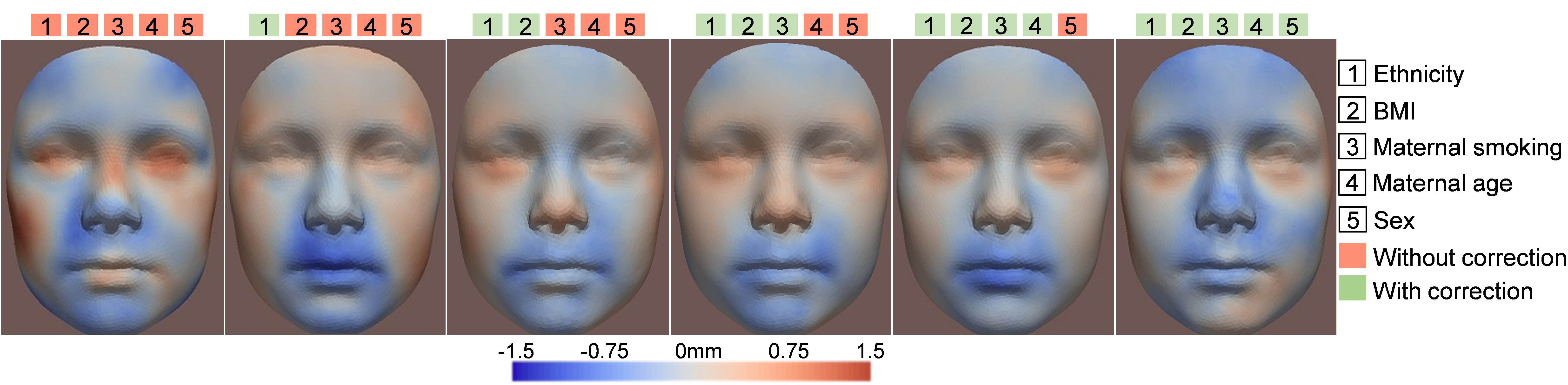}}
\centering
\caption{Interpretation heatmaps of facial changes in children exposed to alcohol during pregnancy (PAE) using the proposed method with gradual correction for confounders of ethnicity, BMI, maternal smoking, maternal age, and sex. From left to right, in the first heatmap no confounder was corrected for during the training; in the last heatmap all five confounders were corrected Equation~\eqref{eq:6_4}. Red areas refer to inward changes of the face with respect to the geometric center of the head.}
\label{fig:fig6_6}
\end{figure}

We provide visual interpretation for the learned association without and with confounder correction (\cref{fig:fig6_5}). In the setting without confounder correction, \cref{fig:fig6_5}a indicates that PAE could lead to a narrow nasal bridge, deep-set eyes, and a narrow cheek. However, such detected facial features might not be the true association with PAE, because they were highly confounded by ethnicity and BMI ($r(\textbf{z}_{\textbf{p}}, \mathrm{ethnicity})=+0.31$ and $r(\textbf{z}_{\textbf{p}}, \mathrm{BMI})=-0.33$ in \cref{tab:tab6_3}). Actually, a narrow nasal bridge and deep-set eyes are common facial features in the Western population \citep{zhang2022genetic,franciscus1991variation}, while a narrow cheek is common in lower-BMI population \citep{jiang2019visual}. After correcting for all confounders, these facial features were not observed in \cref{fig:fig6_5}b anymore. This change suggests that the proposed method successfully removed facial features influenced by the confounders, resulting in a confounder-free association between PAE and children’s facial shape. To provide more insights into the changes caused by each of the confounder variables, we applied the proposed feature interpretation method to visualize our confounder-free results with a gradually increased number of confounders to correct (\cref{fig:fig6_6}, Equation~\eqref{eq:6_4}).

Our results suggest low to moderate maternal alcohol exposure during pregnancy is associated with children’s facial shape. Detected facial phenotypes included turned-up nose tip, shortened nose and turned-in lower-eyelid-related regions. These findings are consistent with previously similar studies \citep{muggli2017association}, and in line with facial abnormality in fetal alcohol spectrum disorders \citep{hoyme2016updated} caused by high levels of PAE.

\subsubsection{Results on global cognition prediction from brain MRI in an elderly population}
\label{sec:result_brain_aging}
\begin{figure}[!h]
\centerline{\includegraphics[trim={0.21cm 0.3cm 2.0cm 1cm}, clip,width=1.0\linewidth]{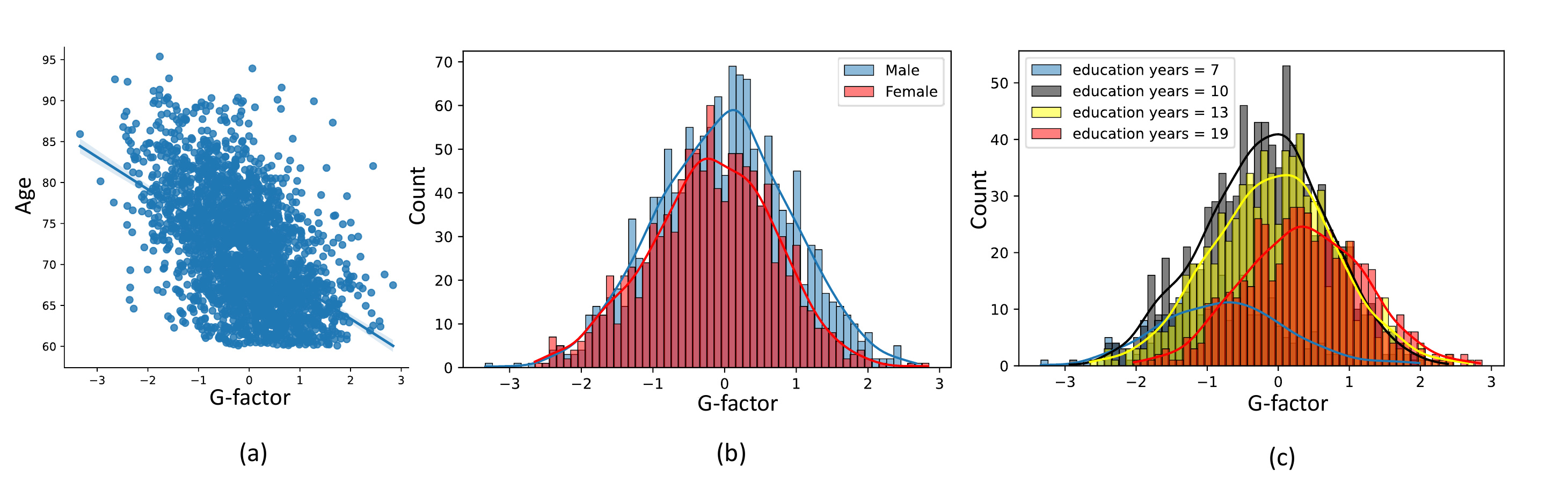}}
\centering
\caption{Data characteristic of the study population. \textbf{(a)} Joint distribution of g-factor and age. The Pearson’s correlation coefficient between age and g-factor is -0.51 (p-value = 4.88e-163, linear regression); \textbf{(b)} Histogram distribution of g-factor between male and female. Male show slightly higher g-factor than female (p-value = 1.4e-5, linear regression); \textbf{(c)} Histogram distribution of g-factor for different educational years. Higher educational years show overall higher g-factors (p-value = 2.87e-57, linear regression)}
\label{fig:fig6_7}
\end{figure}

We applied the proposed method with multiple-confounder correction to study the associations between global cognition and brain imaging in a large elderly population (Rotterdam Study \citep{ikram2020objectives}. As suggested by a previously similar study \citep{roshchupkin2016grey} using traditional statistical method with traditional confounder control, we used g-factor score \citep{hoogendam2014patterns} as learning target variable representing global cognition and used grey matter density maps derived from T1-weighted brain MRIs as input images (Section~\ref{sec:aging_brain_dataset}), and included age, sex (0: male; 1: female), and educational years as confounders. The results show that age is a strong confounder in this study, i.e. participants with an older age generally have a lower g-factor score, (\cref{fig:fig6_7}a, $p<0.001$ via linear regression). The detailed data characteristic of the study population is shown in \cref{fig:fig6_7}. Similar as the Experiment 2, in additional to labelled samples (N=2,395), we included missing-label samples (N=9,406) in the semi-supervised learning. We compared the findings with those obtained without confounder correction.

\begin{table}[!h]
\centering
\caption{Association analysis between the learning target global cognition (g-factor) and brain grey matter imaging. Results are presented without (NA) and with (*) controlling of confounders (age, sex, and educational years).}
\label{tab:tab6_4}
\small 
\setlength{\tabcolsep}{3pt}
\begin{tabular}{l|llll}
\hline
 & Metrics & Ours (NA) & Ours (*) & Ours-SSL (*) \\ \hline
\multirow{2}{*}{\begin{tabular}[c]{@{}l@{}}Prediction \\ performance\end{tabular}} & r-MSE (GF) & \textbf{0.80±0.01} & 0.92±0.02 & 0.91±0.02 \\
 & r(GF) & \textbf{+0.48±0.03} & +0.03±0.03 & +0.05±0.04 \\ \hline
\multirow{4}{*}{\begin{tabular}[c]{@{}l@{}}Confounder\\ removal\end{tabular}} & $r$(Age) & -0.73±0.01 & -0.04±0.03 & \textbf{-0.04±0.03} \\
 & $r$($\mathrm{Age}^2$) & -0.73±0.01 & -0.04±0.03 & \textbf{-0.04±0.03} \\
 & $r$(Sex) & -0.07±0.05 & -0.05±0.02 & \textbf{-0.03±0.02} \\
 & $r$(EY) & +0.13±0.05 & \textbf{+0.03±0.04} & +0.06±0.04 \\ \hline
\multirow{4}{*}{\begin{tabular}[c]{@{}l@{}}Confounder \\ removal\end{tabular}} & $\mathrm{dcor}^2$(Age) & 0.50±0.02 & 0.00±0.00 & \textbf{0.00±0.00} \\
 & $\mathrm{dcor}^2$($\mathrm{Age}^2$) & 0.50±0.02 & 0.00±0.00 & \textbf{0.00±0.00} \\
 & $\mathrm{dcor}^2$(Sex) & 0.01±0.01 & 0.00±0.01 & \textbf{0.00±0.00} \\
 & $\mathrm{dcor}^2$(EY) & 0.02±0.02 & \textbf{0.00±0.00} & 0.00±0.01 \\ \hline
\multirow{4}{*}{\begin{tabular}[c]{@{}l@{}}Confounder\\ removal\end{tabular}} & MI(Age) & 0.39±0.01 & 0.01±0.02 & \textbf{0.00±0.00} \\
 & MI($\mathrm{Age}^2$) & 0.39±0.01 & 0.01±0.02 & \textbf{0.00±0.00} \\
 & MI(Sex) & 0.01±0.01 & \textbf{0.01±0.02} & 0.01±0.02 \\
 & MI(EY) & 0.02±0.02 & 0.01±0.01 & \textbf{0.01±0.01} \\ \hline
\multirow{2}{*}{\begin{tabular}[c]{@{}l@{}}Image\\ reconstruction\end{tabular}} & $L_{1}$-norm & \textbf{0.09±0.00} & 0.09±0.00 & 0.09±0.00 \\
 & NCC & \textbf{0.22±0.00} & 0.22±0.00 & 0.22±0.00 \\ \hline
\multicolumn{5}{p{380pt}}{GF: G-factor.}\\
\multicolumn{5}{p{380pt}}{EY: Educational years.}\\
\multicolumn{5}{p{380pt}}{NCC: Normalized cross correlation.}\\
\multicolumn{5}{p{380pt}}{Ours-SSL (*) means our method with confounder control in a semi-supervised setting utilizing additional missing-label training data.}\\
\multicolumn{5}{p{380pt}}{+ and - indicates a positive or negative correlation between $\textbf{z}_{\textbf{p}}$ and the variables.}\\
\multicolumn{5}{p{380pt}}{A better performance is indicated by a \textbf{bold} value}
\end{tabular}
\end{table}

For the results without correcting for confounders, the correlation coefficient between the learned brain imaging features ($\textbf{z}_{\textbf{p}}$) and g-factor was 0.48 ± 0.03 (\cref{tab:tab6_4}). Even though we explicitly maximized this correlation, it is still lower than the correlation coefficient with age (-0.73 ± 0.01). The observed correlation coefficients for sex and educational years were -0.07 ± 0.05 and +0.13 ± 0.05, lower than that with age. For the results with correction for the confounders of age, sex, and educational years, the correlation coefficient between brain image features ($\textbf{z}_{\textbf{p}}^{*}$) with g-factor decreased to +0.03 ± 0.03, which is much lower than that without confounder correction. On the other hand, the correlation coefficients with the confounders were corrected into a similarly low level, ranged from 0.03 to 0.05. Notably, the image reconstruction quality as measured by the $L_{1}$-norm and normalized cross-correlation (NCC) \citep{li2022ahigh} remained similarly good after removing confounder-related information from brain image features, suggesting that our proposed confounder-free model can still reconstruct the brain morphometry at a high-resolution, with confounder control.

\begin{figure}[!h]
\centerline{\includegraphics[width=1.0\linewidth]{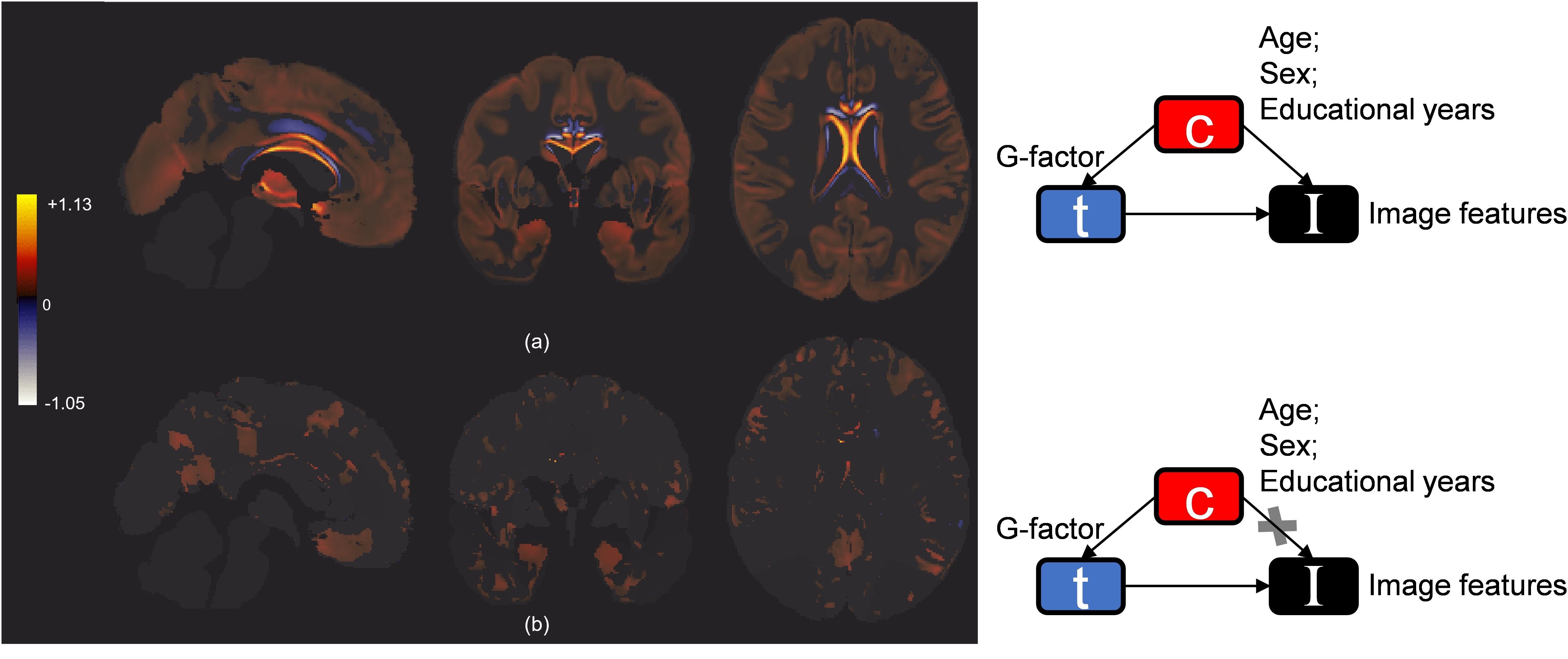}}
\centering
\caption{Reconstructed supratentorial modulated grey matter maps using the sampled latent features along the direction of increasing g-factor \textbf{(a)} without correcting for confounders, and \textbf{(b)} with correcting for age, sex, and educational years. The results are averaged over the five folds, and masked out the statistically non-significant region. Color bar shows the direction and magnitude of the changes of GM density associated with a higher g-factor.}
\label{fig:fig6_8}
\end{figure}

To visualize the anatomical regions linked with global cognition, we reconstructed brain grey matter maps displaying relevant image features (\cref{fig:fig6_8}). Only statistically significant results ($p<0.05$, paired t-test) were visualized, averaging results over five folds. Without confounder correction (\cref{fig:fig6_8}a), the brain heatmaps showed widespread increases in grey matter density with an increasing g-factor, primarily in the cortex, thalamus, and hippocampus. After correcting for the confounding effects of age, sex, and educational years (\cref{fig:fig6_8}b), the highlighted regions were mainly confined to the hippocampus and superior parietal gyrus, indicating a direct association with global cognition. Notably, regions like the left thalamus, initially showing a strong correlation, changed to zero, suggesting a stronger association with confounding variables such as age.

Our findings align with expectations. Similarly, previous studies using traditional statistical methods have demonstrated that education and age represent key neurodevelopmental milestones \citep{lam2013formulation}, after adjusting for key confounders such as age, sex, and educational attainment, the association between brain MRI metrics and cognitive scores remains weak \citep{raz2005regional, raz2010news, roshchupkin2016grey}.

\section{Discussion and conclusion}
In this study, a novel AI method was proposed for conducting association analysis in medical imaging. Our proposed approach effectively addresses the influence of confounding factors by incorporating them as priors, resulting in confounder-free associations. To enhance the interpretability of the outcome associations, a semantic feature visualization approach was proposed, allowing us to gain valuable insights into the image features underlying the observed associations. Moreover, the proposed method supports semi-supervised learning, enabling use of missing-label image data.

The proposed method was applied to two epidemiological association studies. In the second experiment, we analyzed the association between low-moderate prenatal alcohol exposure (PAE) and children’s facial shape after correction for confounders, the proposed method removed facial features related to the confounders (e.g., a narrow cheek or deep-set eyes) and found a remaining correlation of 0.15 between facial features and PAE. In contrast, in the thrid experiment, the analysis of association between brain images and cognitive scores, almost no remaining association (Pearson correlation coefficient $r=$ 0.03 ± 0.03 in \cref{tab:tab6_4}) was found after the correction of confounders. It turned out that the strong association ($r=$ 0.48 ± 0.03 in \cref{tab:tab6_4}) between brain imaging and cognitions before the correction was mainly contributed by the age confounder (\cref{fig:fig6_8}). As these two applications demonstrate, confounder correction is essential as it may prevent wrong or misleading association results. This further highlights the importance of confounder control and model interpretability in AI-based medical image analysis.

The proposed method supports semi-supervised learning (SSL), which has added value in medical image analysis, as for medical image data labels are often missing or may have suboptimal quality. Especially, in cases with only a limited number of labelled samples (say less than 50), SSL improves the image reconstruction quality for feature interpretation as well as the discriminative capacity of the latent features. Moreover, prior work in segmentation \citep{chen2019multi} and classification \citep{gille2023new} supports that SSL by using unlabelled data with reconstruction loss can generally enhance the performance of such joint tasks. 
However, applying this SSL approach with confounder-free prediction as a joint task has not been achieved before due to the conflict between reconstruction loss and confounder removal in prior methods (e.g., VFAE \citep{louizos2015variational}): Confounder removal aims to remove any confounder-related information from the latent space, while reconstruction loss preserves as much information as possible, leading to opposing objectives. This conflict harms the performance of confounder removal when using unlabelled data in SSL. Our method takes a different approach, retaining confounder-related information to strike an optimal balance, thus making this SSL approach possible.
In our experiment results, we only find the improvement of SSL for the facial data but not the brain MRI data  application. The reconstruction quality ($L_{1}$-norm and NCC) was similar between the fully supervised and semi-supervised learning setting. This may be due to the fact that our fully supervised brain autoencoder was optimized with sufficient labeled data, and thus, additional missing-label images could not further improve the optimization of the brain autoencoder.

A limitation of the proposed method is that it requires human prior knowledge for the identification of confounders. In the future, we will consider integrating techniques from causal inference \citep{gao2015local} with the proposed method for the automatic identification of potential confounders. In addition, a future direction to explore is how to incorporate input data with a discrete distribution, since our reconstruction-based feature interpretation technique presumes a continuous latent space. One possible way is using a variational autoencoders, which enforces a Gaussian distribution in the latent space.

In conclusion, our AI method, complemented by its semi-supervised variant, offers a promising toolset for enhancing association analysis in medical imaging. Future research can further refine and extend this method, ensuring more robust and interpretable findings in medical imaging studies.

\section*{CRediT authorship contribution statement}
All authors made significant contributions to this scientific work and approved the final version of the
manuscript. X.L. and B.L. were involved in the conception and design of the study, conducted the
method development and experiments, and wrote the article. E.E.B. and G.V.R. were involved in the
conception and design of the study, supervised the method development and design of experiments,
and co-wrote the article. M.W.V. and E.B.W. were involved in the conception and design of the study,
reviewed the manuscript, and provided consultation regarding the analysis and interpretation of the
data.

\section*{Declaration of competing interest}
The authors declare that they have no known competing financial interests or personal relationships that could have appeared to
influence the work reported in this paper.

\section*{Data availability}
Code and synthetic image data are available at: \url{https://gitlab.com/radiology/compopbio/ai\_based\_association\_analysis}

\section*{Acknowledgments}
The Generation R Study is conducted by the Erasmus MC in close collaboration with the School of Law
and Faculty of Social Sciences of the Erasmus University Rotterdam, the Municipal Health Service
Rotterdam area, Rotterdam, the Rotterdam Homecare Foundation, Rotterdam, and the Stichting
Trombosedienst \& Artsenlaboratorium Rijnmond (STAR-MDC), Rotterdam. We gratefully acknowledge
the contribution of children and parents, general practitioners, hospitals, midwives, and pharmacies in
Rotterdam.

The Rotterdam Study is funded by the Erasmus Medical Center and Erasmus University, Rotterdam,
Netherlands Organization for the Health Research and Development (ZonMw), the Research Institute for
Diseases in the Elderly, the Ministry of Education, Culture and Science, the Ministry for Health, Welfare
and Sports, the European Commission (Directorate-General XII), and the Municipality of Rotterdam. The
authors are grateful to the study participants, the staff from the Rotterdam Study, and the participating
general practitioners and pharmacists.

\bibliographystyle{model2-names.bst}\biboptions{authoryear}
\bibliography{refs}

\appendix
\section{}
\label{appendix6a1}

\subsection{Details of the extended synthetic ellipse image data set}
\begin{figure}[!h]
\centering\includegraphics[width=0.85\linewidth]{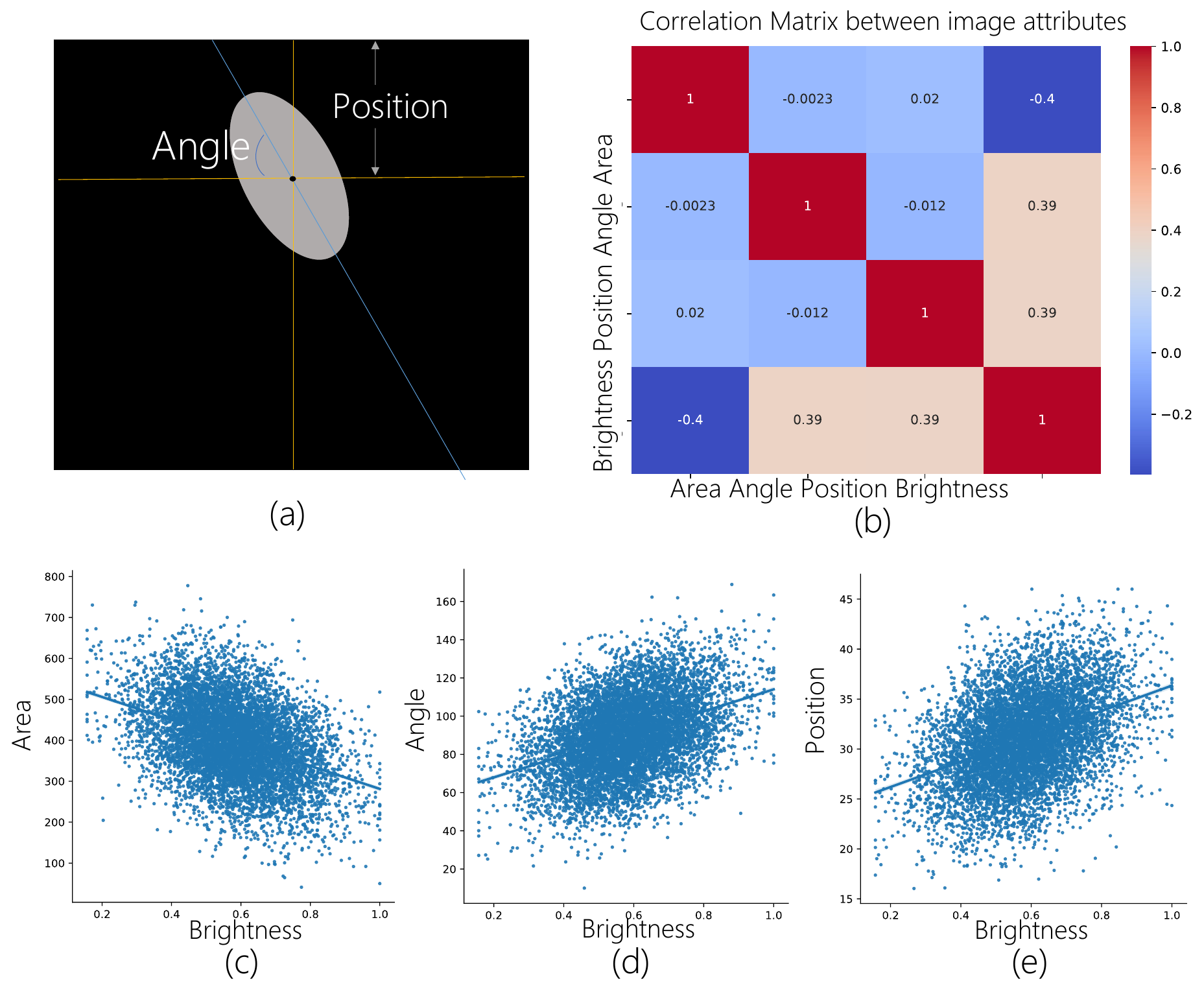}
\caption{Characteristics of the extended synthetic dataset. (a) Illustration of the angle and position definitions for an ellipse.
(b) Correlation matrix of the four image-related attributes.
(c) Joint Gaussian distribution between brightness and area, showing that brighter ellipses tend to have smaller sizes.
(d) Joint Gaussian distribution between brightness and angle, showing that brighter ellipses tend to rotate clockwise.
(e) Joint Gaussian distribution between brightness and position, showing that brighter ellipses tend to shift toward the bottom of the image.}
\label{fig:fig6_a_1}
\end{figure}

To further demonstrate the effectiveness of the proposed method in dealing with multiple confounders, we extended the synthetic dataset
into an ellipse dataset (N=8,000; image size 64x64), featuring four image-related attributes: brightness, area, rotation angle, and center
position (\cref{fig:fig6_a_1}a). Each attribute follows a Gaussian distribution, with detailed characteristics provided in \cref{tab:tab6_a_1}. All ellipses have the same height-to-width ratio, allowing the area to visually reflect their size.

We used brightness as the learning target, while area, position, and angle were treated as confounders. To simulate confounding effects,
we generated a multivariate Gaussian distribution between brightness and each confounder (with a Pearson’s correlation of 0.4), while
ensuring independence among the confounders (Pearson’s correlation: 0). \cref{fig:fig6_a_1}b illustrates the correlations between the four image-related attributes.

\begin{table}[!h]

\centering
\caption{Characteristic of the four image-related attributes.}
\label{tab:tab6_a_1}
\setlength{\tabcolsep}{3pt}

\begin{tabular}{l|llll}
\hline
 & Mean (SD) & Range & Representation & Note \\ \hline
Brightness & 0.58 (0.14) & 0.16-1.00 & Gray level 40-255 & Target, t \\
Angle & 90 (20.33) & 10 -169 degree & Clockwise   rotation & Confounder, c1 \\
Position & 31 (4.50) & 16-46 pixel & Vertical translation & Confounder c2 \\
Area & 399 (97.90) & 41-778 pixels & Size of ellipses & Confounder c3 \\ \hline
\end{tabular}
\end{table}

\newpage
\subsection{Experiments and results of the ellipse dataset}
\begin{figure}[!h]
\centering\includegraphics[width=0.8\linewidth]{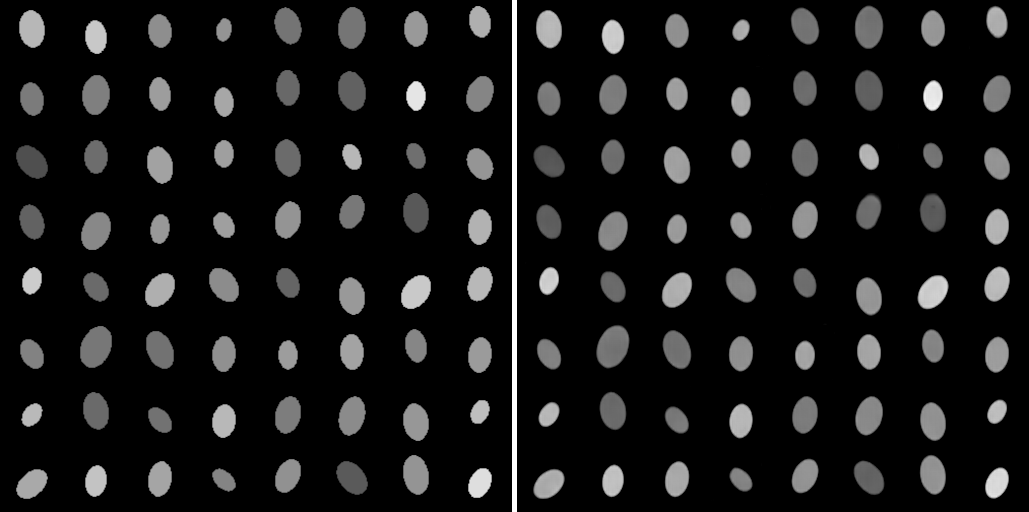}
\caption{The input images (left, 8x8 ellipse images) and reconstructed results (right).}
\label{fig:fig6_a_2}
\end{figure}

We used the same experiment settings as the original synthetic circle dataset described in the manuscript,
with the only modification being an increase in the latent dimension, from $n$=2 to $n$=8, to accommodate this more
complex scenario. Notably, in case of $m$ confounders, our method requires a latent dimension $n \geq m + 1$ to guarantee
there exist a vector $\vec{{p}^{*}}$ orthogonal to the $m$ confounder-related vectors in the latent space. \cref{fig:fig6_a_2} presents the
input and reconstructed images after model training.

\begin{figure}[!h]
\centering\includegraphics[width=1.0\linewidth]{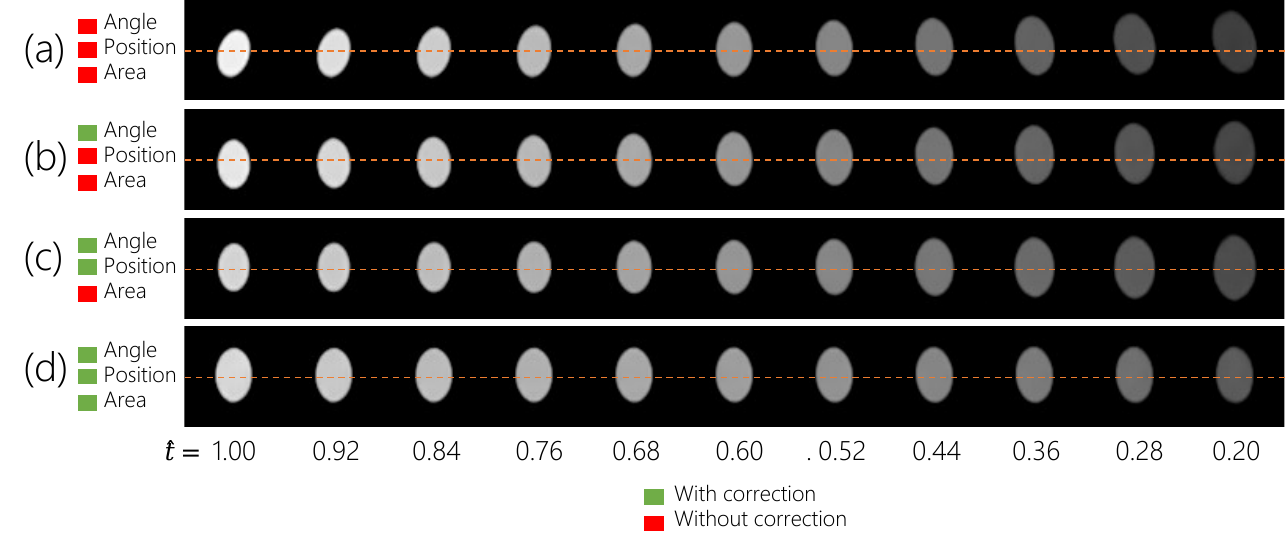}
\caption{Semantic feature visualization of our method for the ellipse experiment.
a) No confounders were included in Equation~\eqref{eq:6_4} during model training; b) The confounder ‘angle’ was included in Equation~\eqref{eq:6_4};
c) The confounders ‘angle’ and ‘position’ were included in Equation~\eqref{eq:6_4}; d) The confounders ‘angle’, ‘position’,
and ‘area’ were included in Equation~\eqref{eq:6_4}. The sampling scale of brightness was normalized to 0.2-1.0 using
\textbf{Algorithm~\ref{alg:semantic_feature_visualization}} for all experiments. We manually added the dash lines to enhance visibility.}
\label{fig:fig6_a_3}
\end{figure}

To further demonstrate the effectiveness of the proposed method in dealing with multiple confounders, and to provide deeper insights
into the image features associated with each confounder, we applied the proposed feature interpretation method, while gradually
increasing the number of confounders to correct. \cref{fig:fig6_a_3} presents the qualitative results, and \cref{tab:tab6_a_2} provides the quantitative results.

In \cref{fig:fig6_a_3}a, without correction for any confounder, the reconstructed images display decreasing brightness ($r_{brightness}$=-0.997,
as shown in the column ‘Ours (NA)’ in \cref{tab:tab6_a_2}). However, they also exhibited counterclockwise rotation ($r_{angle}$=-0.391),
upwards translation ($r_{position}$=-0.386), and increasing size ($r_{area}$=+0.397);\\
In \cref{{fig:fig6_a_3}}b, after incorporating the confounding variable ‘angle’ into Equation~\eqref{eq:6_4}, the rotation effects of were removed in reconstructed images;\\
In \cref{{fig:fig6_a_3}}c, the translation effects were further eliminated after correction;\\
In \cref{{fig:fig6_a_3}}d, the effects of all confounders were removed, leaving only the brightness effects in the reconstructed images. Correspondingly,
r(angle), r(position), and r(area) are reduced to -0.021, -0.028 and +0.035, respectively, as shown in the last column of \cref{tab:tab6_a_2}.

\begin{table}
\begingroup
\tiny 
\caption{Quantitative results of the ellipse experiment.}
\label{tab:tab6_a_2}

\begin{tabular}{l|l|llll}
\hline
 & Metrics & \begin{tabular}[c]{@{}l@{}}Ours \\ (NA)\end{tabular} & \begin{tabular}[c]{@{}l@{}}Ours \\ (*Angle)\end{tabular} & \begin{tabular}[c]{@{}l@{}}Ours \\ (*Angle,\\    \\ Pos)\end{tabular} & \begin{tabular}[c]{@{}l@{}}Ours \\    \\ (*Angle,\\    \\ Position,\\    \\ Area)\end{tabular} \\ \hline
\multirow{2}{*}{\begin{tabular}[c]{@{}l@{}}Prediction \\ performance\end{tabular}} & r-MSE & \textbf{0.013+-0.001} & 0.058±0.001 & 0.080±0.001 & 0.098±0.002 \\
 & r(brightness) & \textbf{-0.997+-0.000} & -0.907±0.006 & -0.814±0.005 & -0.704±0.011 \\ \hline
\multirow{9}{*}{\begin{tabular}[c]{@{}l@{}}Confounder \\ removal\end{tabular}} & r(angle) & -0.391+-0.027 & -0.021±0.026 & -0.029±0.020 & \textbf{-0.021±0.022} \\
 & r(position) & -0.386+-0.017 & -0.421±0.028 & \textbf{-0.020±0.023} & -0.028±0.017 \\
 & r(area) & +0.397+-0.020 & +0.423±0.019 & +0.453±0.007 & \textbf{+0.035±0.023} \\
 & $\mathrm{dcor}^2$(angle) & 0.121+-0.015 & 0.006±0.002 & 0.006±0.001 & \textbf{0.005±0.001} \\
 & $\mathrm{dcor}^2$(position) & 0.123+-0.011 & 0.148±0.019 & \textbf{0.004±0.001} & 0.006±0.001 \\
 & $\mathrm{dcor}^2$(area) & 0.128+-0.017 & 0.147±0.017 & 0.173±0.008 & \textbf{0.004±0.002} \\
 & MI(angle) & 0.084+-0.018 & 0.035±0.029 & 0.046±0.016 & \textbf{0.030±0.016} \\
 & MI(position) & 0.086+-0.008 & 0.101±0.018 & 0.044±0.017 & \textbf{0.041±0.028} \\
 & MI(area) & 0.105+-0.012 & 0.108±0.015 & 0.125±0.012 & \textbf{0.032±0.015} \\ \hline
\begin{tabular}[c]{@{}l@{}}Image \\ reconstruction\end{tabular} & L1-norm & 0.007+-0.000 & \textbf{0.006±0.000} & 0.007±0.001 & -0.007±0.001 \\ \hline
\multicolumn{6}{p{350pt}}{Ours (NA): No confounders were included in Eq. 4 during model training.}\\
\multicolumn{6}{p{350pt}}{Ours (*Angle): The confounder ‘angle’ was included in Equation~\eqref{eq:6_4}.}\\
\multicolumn{6}{p{350pt}}{Ours (*Angle, Position): The confounder ‘Angle’ and ‘Position’ were included in Equation~\eqref{eq:6_4}.}\\
\multicolumn{6}{p{350pt}}{Ours (*Angle, Position, Area): All defined confounders were included in Equation~\eqref{eq:6_4}.}\\
\multicolumn{6}{p{350pt}}{A better performance is indicated by a \textbf{bold} value.}
\end{tabular}
\endgroup
\end{table}

These results further confirm the effectiveness of the proposed method in mitigating the confounding effects of multiple confounders.
In addition, based on the geometric insights in \cref{{fig:fig6_1}}a, the theoretical upper bound of remaining correlation between zp and t can be estimated as
$|r(\textbf{z}_{\textbf{p}}^{*},  \textbf{t})| \leq \sqrt{1-r^{2}(\textbf{t}, \textbf{c}_{\textbf{1}})-r^{2}(\textbf{t}, \textbf{c}_{\textbf{2}})-r^{2}(\textbf{t}, \textbf{c}_{\textbf{3}})}=0.721$, subject to  $r(\textbf{z}_{\textbf{p}}^{*},  \textbf{c}_{\textbf{1}})= r(\textbf{z}_{\textbf{p}}^{*},  \textbf{c}_{\textbf{2}})= r(\textbf{z}_{\textbf{p}}^{*},  \textbf{c}_{\textbf{3}})=0$
(Extension from Equation~\eqref{eq:6_1}, as $c_1$, $c_2$, $c_3$ are independent with each other).
This is in line with our experimental results $|r(brightness)|$=0.704, when controlling confounders at r(angle)= -0.021, r(position)= -0.028, r(area)= +0.035 (\cref{tab:tab6_a_2}).

We uploaded the ellipse dataset and scripts to our Gitlab repository to enhance the reproducibility of our method and the results.

\clearpage
\subsection{Geometry interpretation of multiple confounders in the latent space.}

\begin{figure}[!h]
\centering\includegraphics[width=1.0\linewidth]{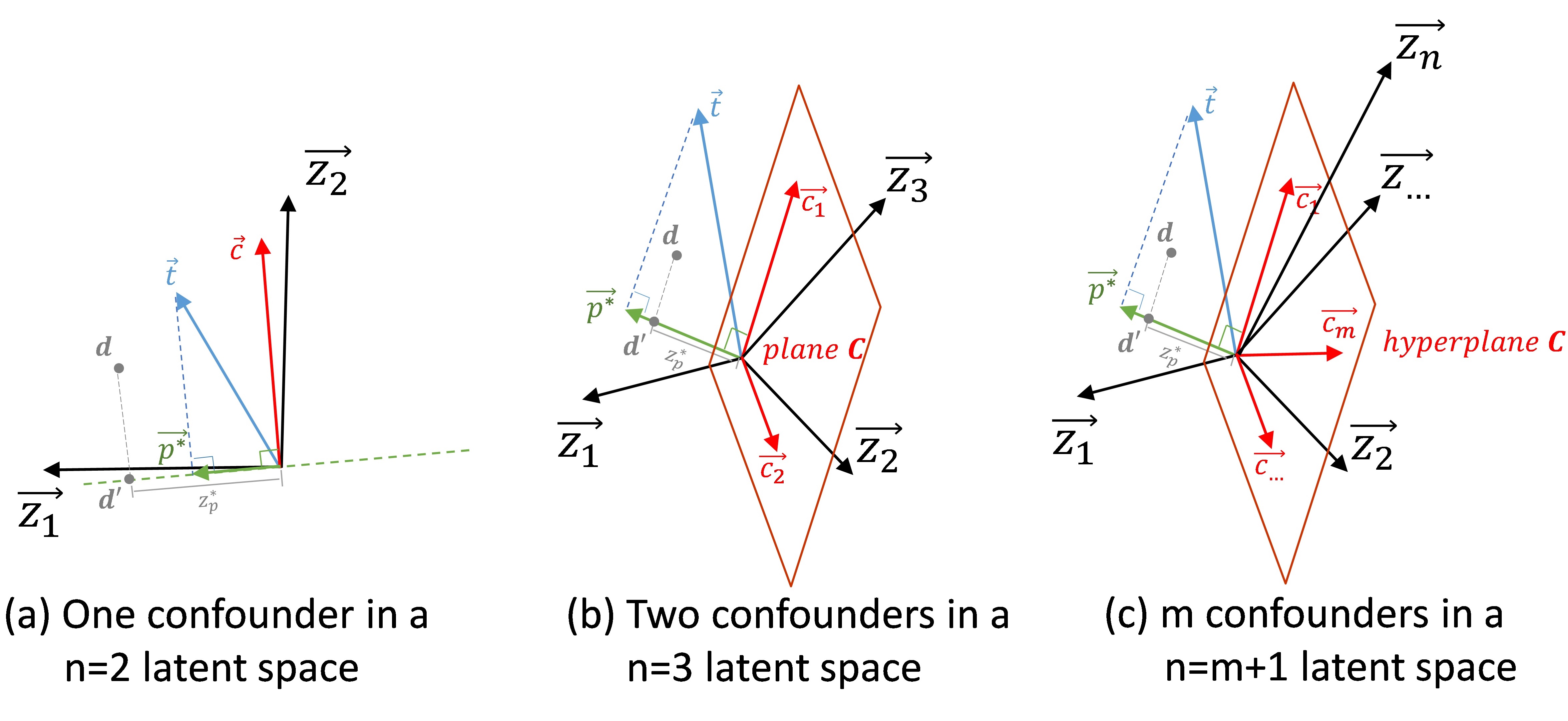}
\caption{Geometry interpretation of correlations between a target variable and $m$ confounder variables in the latent space,
where the linear correlation between two variables (e.g., ${z}_{p}^{*}$, $c$) is represented by the cosine of the angle between their corresponding vectors
(e.g., $\vec{{p}^{*}}$, $\vec{c}$). a) A single confounder ($m=1$) in a 2-dimensional latent space ($n=2$), where vector $\vec{{p}^{*}}$ is orthogonal to vector $\vec{c}$;
b) Two confounders ($m=2$) in a 3-dimensional latent space ($n=3$), where vector $\vec{{p}^{*}}$ is orthogonal to vector $\vec{c_1}$ and $\vec{c_2}$;
c) $m$ confounders in an $n=m+1$ latent space, where vector $\vec{{p}^{*}}$ is orthogonal to all $m$ confounders ${\vec{c_1},\vec{c_2},...,\vec{c_m}}$.}
\label{fig:fig6_a_4}
\end{figure}


\end{document}